\documentclass[pdflatex,sn-mathphys-num,iicol]{sn-jnl}

\usepackage{graphicx}%
\usepackage{multirow}%
\usepackage{amsmath,amssymb,amsfonts}%
\usepackage{amsthm}%
\usepackage{mathrsfs}%
\usepackage{xcolor}%
\usepackage{textcomp}%
\usepackage{manyfoot}%
\usepackage{booktabs}%
\usepackage{algorithm}%
\usepackage{algorithmicx}%
\usepackage{algpseudocode}%
\usepackage{listings}%
\usepackage{subcaption}
\usepackage{stfloats}
\usepackage{enumitem}
\usepackage[T1]{fontenc}
\usepackage{lmodern}
\usepackage{todonotes}
\usepackage{changes}
\usepackage{wrapfig}
\usepackage[protrusion=true,expansion=true]{microtype}
\usepackage{hyperref}
\usepackage{mdframed}
\usepackage{ulem}

\hypersetup{
  pdftitle={Do Robots Really Need Anthropomorphic Hands? A Comparison of Human and Robotic Hands},
  pdfauthor={Fabisch, Zai El Amri, Singh, and Navarro-Guerrero},
  pdfsubject={Robotics (cs.RO); Artificial Intelligence (cs.AI)},
  pdfkeywords={Object Manipulation, Robotics Hands, Grippers, Smart Prosthetics, Manipulation, Grasping, Survey, Review},
  colorlinks=true,
  linktoc=all,
  linkcolor=darkgray,
  citecolor=darkgray,
  urlcolor=black,
}

\begin{document}

\title[Do Robots Really Need Anthropomorphic Hands?]{Do Robots Really Need Anthropomorphic Hands? A Comparison of Human and Robotic Hands}

\author[1]{\fnm{Alexander} \sur{Fabisch}}\email{alexander.fabisch@dfki.de}\equalcont{These authors contributed equally to this work.}

\author*[2]{\fnm{Wadhah} \sur{Zai El Amri}}\email{wadhah.zai@l3s.de}\equalcont{These authors contributed equally to this work.}

\author[1]{\fnm{Chandandeep} \sur{Singh}}\email{singhchandandeep34@gmail.com}

\author[2]{\fnm{Nicol\'as} \sur{Navarro-Guerrero}}\email{nicolas.navarro.guerrero@gmail.com}

\affil[1]{\orgdiv{Robotics Innovation Center}, \orgname{Deutsches Forschungszentrum f\"ur K\"unstliche Intelligenz GmbH (DFKI)}, \orgaddress{\street{Robert-Hooke-Straße 1}, \city{Bremen}, \postcode{28359}, \state{Bremen}, \country{Germany}}}

\affil*[2]{\orgname{Leibniz Universität Hannover}, \orgdiv{\href{https://www.l3s.de/}{L3S Research Center}}, \orgaddress{\street{Appelstra\ss e}, \city{Hanover}, \postcode{30167}, \state{Lower Saxony}, \country{Germany}}}

\abstract{%
Human manipulation skills represent a pinnacle of voluntary motor functions, requiring the coordination of many degrees of freedom and the processing of high-dimensional sensor input to achieve remarkable dexterity.
Thus, this study investigates whether the human hand, with its associated biomechanical properties, sensors, and control mechanisms, is an ideal that should be strived for in robotics. Do robots need anthropomorphic hands?

First, characteristics of the human hand in terms of biomechanics and perception are extracted to compare them with currently commercially available robotic hands.
From this comparison, research questions are derived that connect manipulation system complexity to skill repertoire size and dexterity.
These questions are addressed through a systematic literature review, analyzing the manipulation capabilities demonstrated in 125 papers published between 2019 and 2025.

Although complex five-fingered hands are often considered the ultimate goal for robotic manipulators, they are not necessary for all tasks.
Findings indicate that in-hand manipulation does not benefit from anthropomorphic hand design, as simpler mechanisms are sufficient; however, mechanism complexity correlates with the breadth of manipulation tasks a hand can perform.
Sensor integration and intelligent manipulation strategies remain underexplored, which may be due to a misalignment with hand design: instead of replicating the number of fingers and degrees of freedom, focusing on robustness and softness would allow more intelligent control and learning to exploit environmental contacts and integrate more sensors.
Finally, the article argues for standardized evaluation criteria to enable the systematic comparison of hand designs and manipulation systems.
}


\maketitle

\section{Introduction}
\begin{figure*}[bt]
\centering
\begin{subfigure}[c]{0.3\textwidth}
\centering
\includegraphics[width=\textwidth]{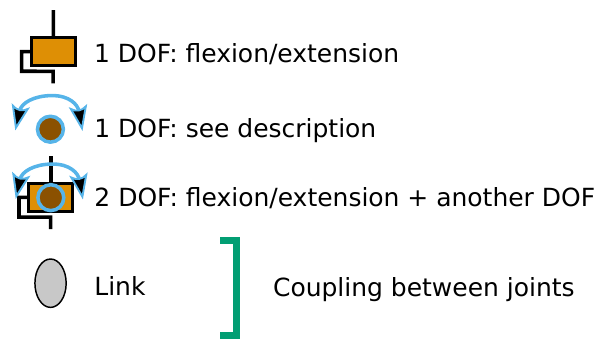}
\caption{Legend for kinematic diagrams.}
\label{fig:kin_legend_fig1}
\end{subfigure}
\hfill
\begin{subfigure}[c]{0.6\textwidth}
\centering
\includegraphics[width=\textwidth]{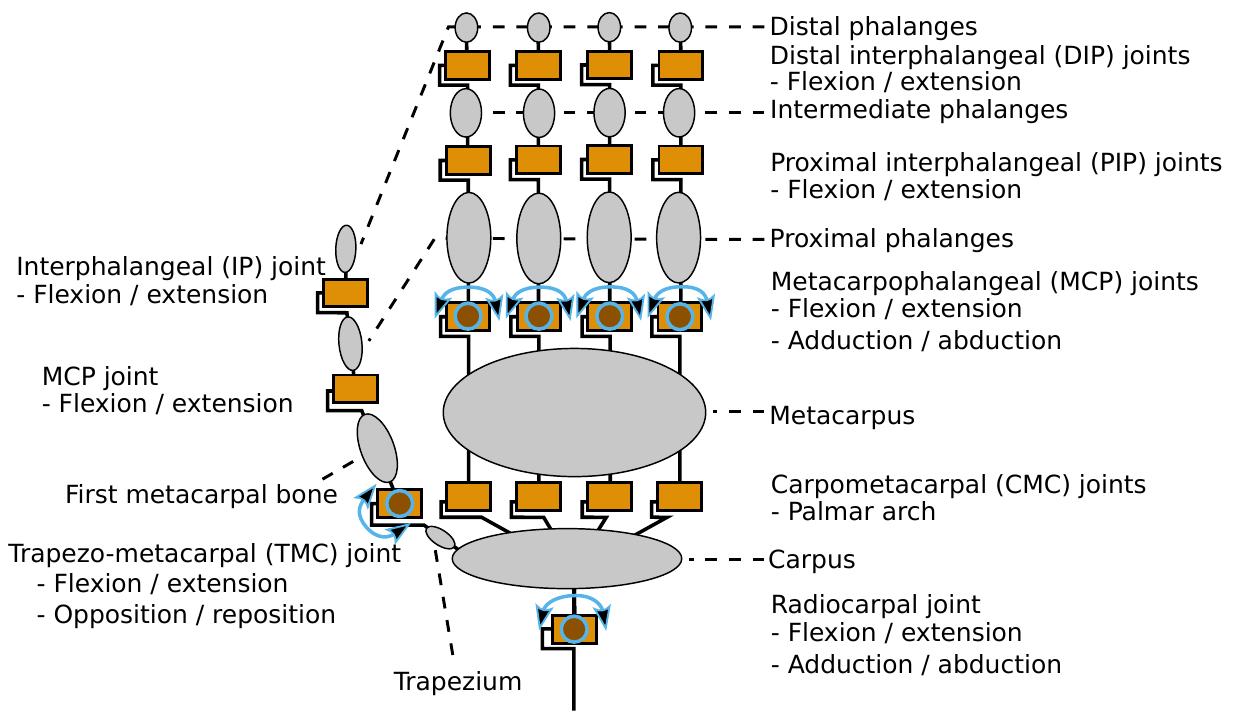}
\caption{Kinematics of the human hand.}
\end{subfigure}
\caption{Kinematics of the human hand. All fingers have four degrees of freedom. The thumb is opposable with its saddle joint. All other fingers support abduction/adduction. All fingers support flexion/extension in three joints. Notation based on Wang et al.~\cite{Wang2008Topology}.}
\label{fig:kin_human_hand}
\end{figure*}

The human hand (kinematic diagram in Fig.~\ref{fig:kin_human_hand}) is often used as the gold standard for robotic manipulation. Anthropomorphic robots that are capable of complex manipulation skills are often used to showcase the frontier of robotics and artificial intelligence \cite{OpenAI2019Solving}.

Despite our fascination for anthropomorphic robotic hands, robots in industrial applications and robotics competitions use simpler designs such as parallel grippers or suction cups \cite{Eppner2018Four,Corbato2018Integrating,Billard2019Trends}.
This is not necessarily only due to the complexity of matching the human hand's dexterity, robustness, and perceptual capabilities, but perhaps also to the reality that anthropomorphic hands might not be required to manipulate efficiently in all cases \cite{Bicchi2000Hands}.
For instance, the first winner of the Amazon Picking Challenge used an end effector based on a suction system \cite{Eppner2018Four}. In the DARPA Robotics Challenge, 15 of 25 teams used an underactuated hand with three or four fingers, while none of the remaining ten teams used a fully actuated anthropomorphic hand \cite{Piazza2019Century}.
The same pattern holds in assistive technology: the winner of the Cybathlon\footnote{Cybathlon is a global challenge organized by ETH Zurich to develop assistive technologies suitable for everyday use for people with disabilities.} Powered Arm Prosthesis Race used a body-powered hook~\cite{Piazza2019Century}.
Furthermore, birds are often able to perform complex manipulations, such as nest building, just with their beaks \cite{sugasawa2021object}.

Yet humans themselves often need both hands for complex tasks such as tying shoe laces, suggesting that one opposable thumb and five fingers may be insufficient.
This is supported by the fact that individuals with six fully functional fingers per hand can tie shoe laces with one hand and achieve superior manipulation abilities in other tasks~\cite{Mehring2019Augmented}.
When designing new robotic hands from scratch, we could improve the dexterity, flexibility, and versatility of the mechanism, for instance, by introducing two opposable thumbs or six fingers~\cite{Billard2019Trends}.

These observations prompt the question of whether we need anthropomorphic robotic hands. To answer it, we (1) characterize the human hand's kinematics and perception, (2) survey commercially available robotic and prosthetic hands, and (3) perform a systematic literature review to correlate hand features with demonstrated skills and in-hand manipulation complexity.

Related surveys focus on artificial intelligence to achieve human-like dexterous manipulation \cite{Huang2025Human,Li2026Developments}, hardware design and control of dexterous robotic hands \cite{Zhao2026Robotic}, and describe trends in robotic hand research \cite{Piazza2019Century,Billard2019Trends}.
Only a few authors question the usefulness of anthropomorphic hand design (e.g., \cite{Billard2019Trends}) and none addresses this topic with a systematic review.

\section{The Human Hand}
The human hand is one of the most advanced manipulation devices that can be found in nature~\cite{Kapandji2007Physiology}.
To answer whether robotic hands need to be more like human hands, we define the characteristics of the human hand in terms of biomechanics, sensors, control, and learning.

\subsection{Biomechanics}
\label{sec:humanHand_biomechanics}

The human hand usually has five fingers, an opposable thumb, a high number of about 24 degrees of freedom (DoF)~\cite{Cobos2008Efficient}, including abduction/adduction and three flexion/extension joints per finger, of which most of them are almost individually controllable~\cite{Hager-Ross2000Quantifying}, an extensive active and even larger passive range of motion, a soft and deformable surface, particularly at the large surface of the flexible palm, and the ability to produce high grip forces.
However, individual control of each degree of freedom is not perfect for several reasons~\cite{Slobounov2002Role,Appell2008Funktionelle,Hager-Ross2000Quantifying}. For precision grasps with small contact areas, abduction/adduction of fingers 2--5 seems to be more relevant than individual flexion/extension to oppose the thumb~\cite{Montagnani2016Independent}.
Dexterous manipulation relies on a flexible arm with seven DoF to which the hand is attached~\cite{Montagnani2015it}.
The exact dimensions and ranges of motions vary from person to person.

\begin{figure*}[tb]
\centering
\includegraphics[width=0.8\textwidth]{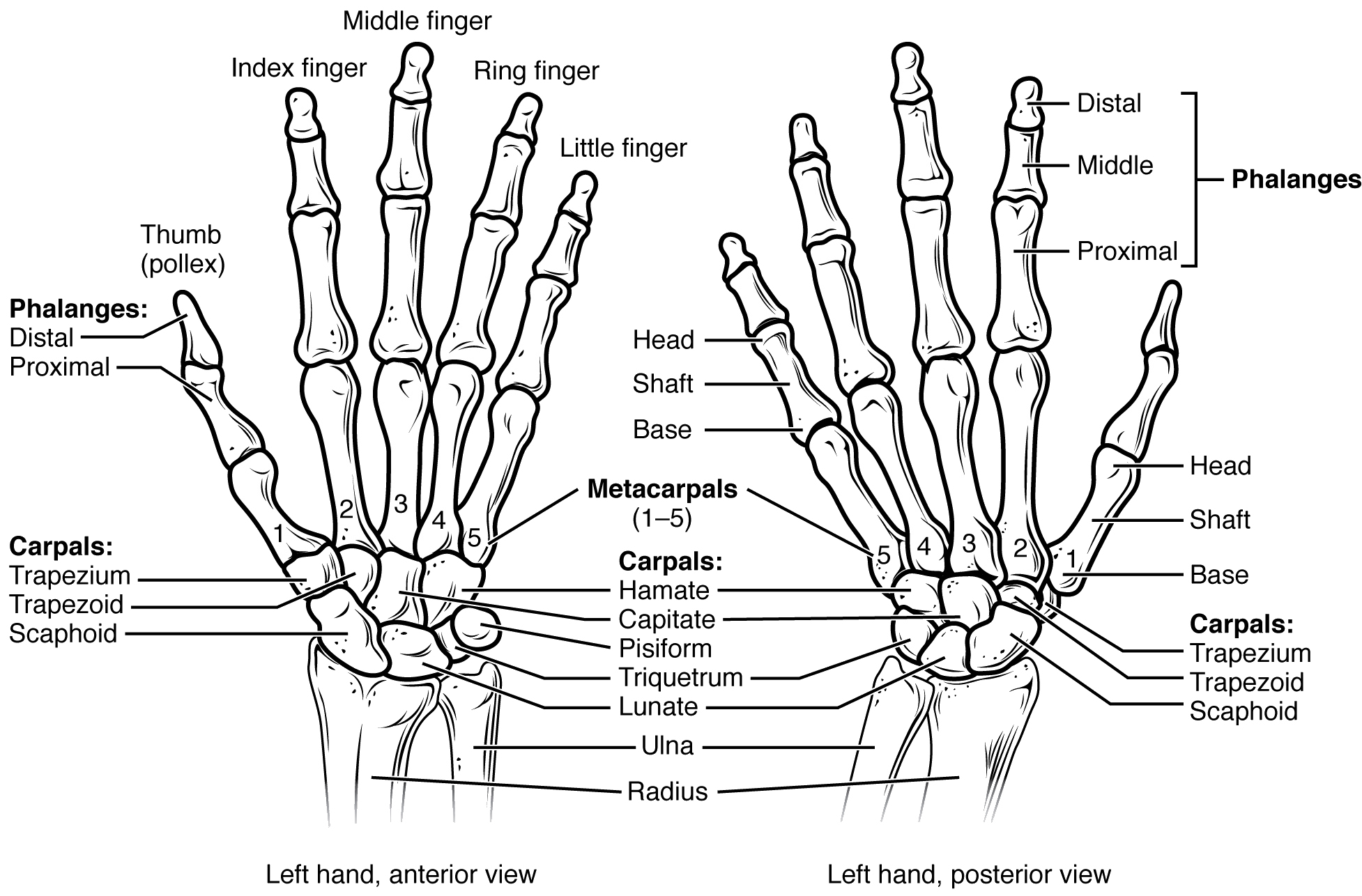}
\caption{Bones of the human hand. Image from \cite{Young2013Anatomy} CC BY 4.0.}
\label{fig:bones-hand}
\end{figure*}

\subsubsection{Fingers}
Fingers are enumerated from thumb (1) to little finger (5).
Although most human hands have five fingers (see Fig.~\ref{fig:bones-hand}), their number can vary, as some people have six fully functional fingers~\cite{Mehring2019Augmented}.

It is not clear, however, how many fingers are necessary.

Many human grasps need only the thumb and another virtual finger \cite{Montagnani2016Independent}. More degrees of freedom (DoF) in the human hand are mostly necessary for precision grasps, which are used less often.
 Only three fingers exert force in most human grasp types of a human grasp taxonomy~\cite{Feix2016GRASP} and other fingers are only used to fine-tune the grasp~\cite{Abbasi2016Grasp}.
Two or three fingers in contact with an object increase workspace volume and rotational range when compared to more fingers in the specific case of in-hand precision manipulation of circular objects with the human hand~\cite{Feix2021Effect}.
In contrast to these results, individuals with six fully functional fingers can solve the task of tying shoe laces with only one hand, whereas most individuals with five fingers would need two hands~\cite{Mehring2019Augmented}. Hence, more fingers may increase manipulation capabilities.

\subsubsection{Joints and Degrees of Freedom}
Not all of its many bones (see Fig.~\ref{fig:bones-hand}) are connected through movable joints, but the human hand has at least 20 DoF---four in each finger. Fingers 2--5 perform flexion/extension in three joints\footnote{MCP: metacarpo-phalangeal joint; PIP: proximal interphalangeal joint; DIP: distal interphalangeal joint.} and abduction/adduction\footnote{Abduction/adduction: moving away from / closer to the middle finger.} in one joint (MCP). The thumb has a saddle joint (TMC: trapezo-metacarpal joint) with two DoF (flexion/extension and opposition/reposition), and two more joints (MCP; IP: interphalangeal joint) for flexion/extension. It is possible to move the metacarpal bones and hollow the palm to a limited extent~\cite{Kapandji2007Physiology}. Hence, Cobos~et~al.\ \cite{Cobos2008Efficient} define 24 DoF. Assuming the carpus to be rigid is an approximation though~\cite{Kapandji2007Physiology}, and the palm, as well as the skin, are soft and deformable. A kinematic diagram is displayed in Fig.~\ref{fig:kin_human_hand}.

Cobos~et~al.\ \cite[Table III]{Cobos2008Efficient} provide a concise overview of the ranges of motion.\footnote{Range of flexion seems to be swapped between MCP and PIP joints~\cite{Kapandji2007Physiology}.} We distinguish an active range of motion reachable through muscle contraction and a larger passive range of motion only reachable with external force (e.g., DIP extension: 5° vs. 30°~\cite{Kapandji2007Physiology}).
The exact ranges vary. For example, distal hyperextensibility of the thumb allows to extend the thumb's IP joint by 90°, resulting in a 180° range.

\subsubsection{Muscles and Tendons}

\begin{figure}[b!]
\centering
\begin{subfigure}{\columnwidth}
\includegraphics[width=\textwidth]{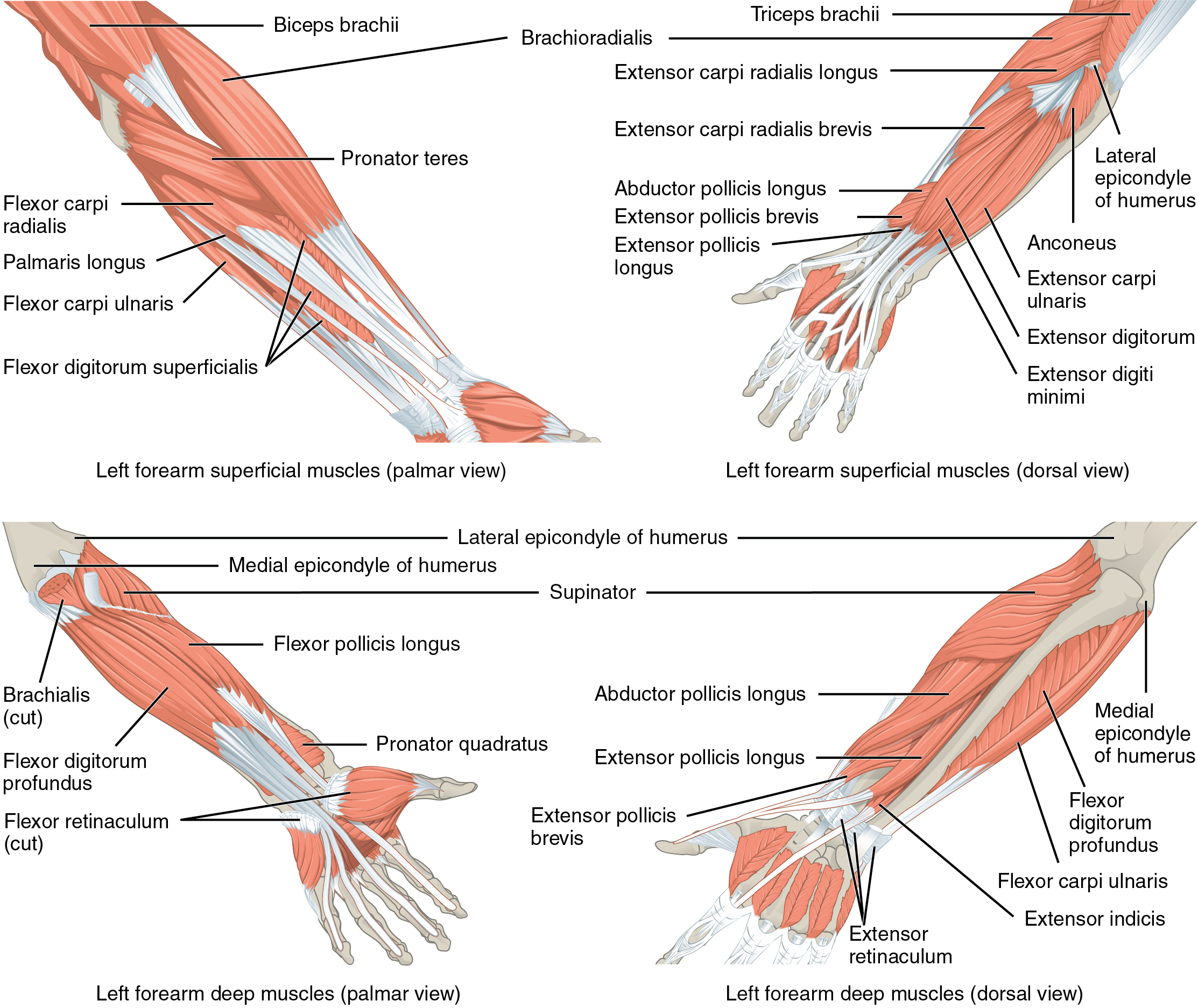}
\caption{Muscles in the forearm move wrist and fingers.}
\label{fig:muscles-forearm}
\end{subfigure}
\begin{subfigure}{\columnwidth}
\centering
\includegraphics[width=\textwidth]{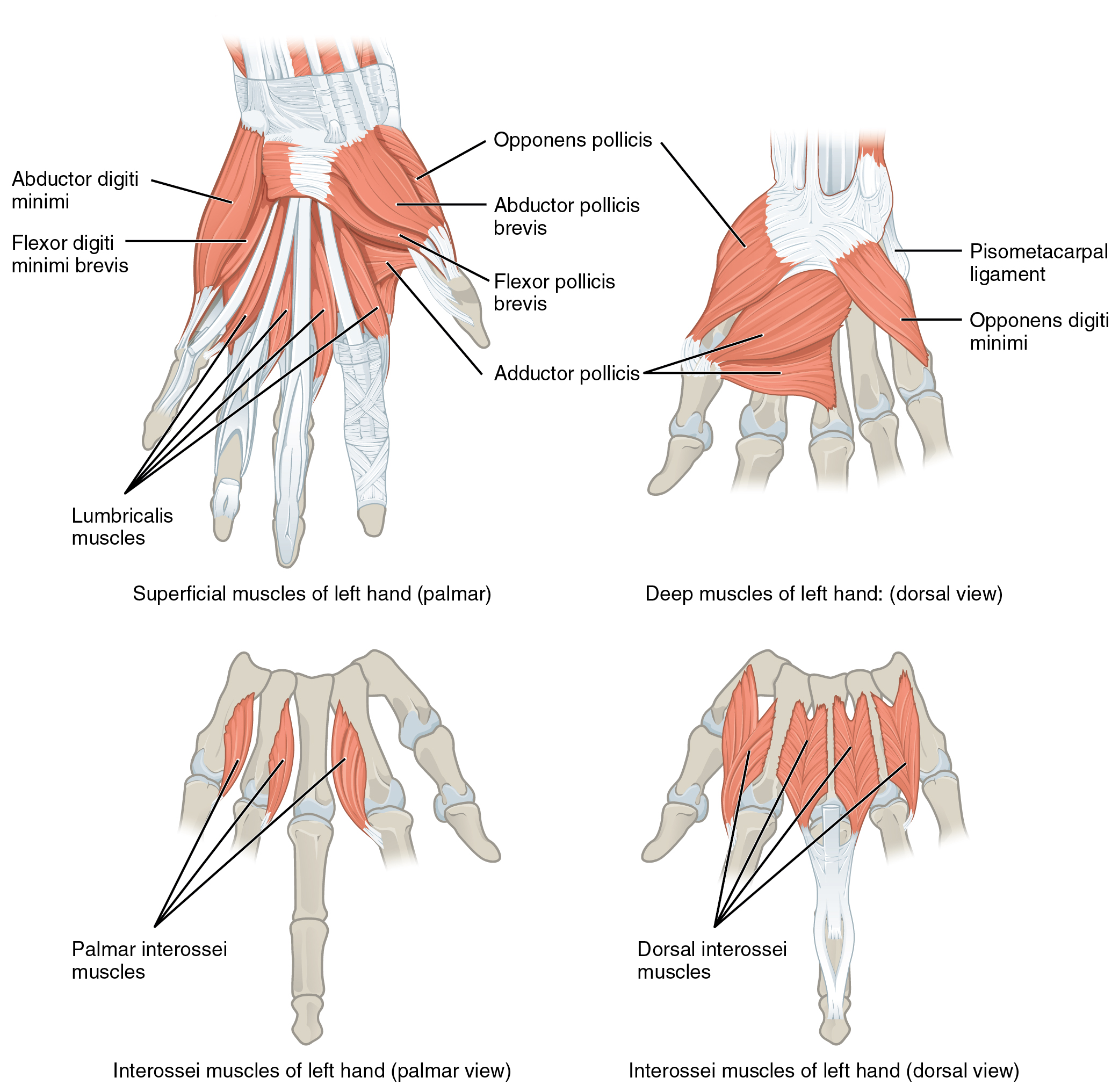}
\caption{Intrinsic muscles of the human hand.}
\label{fig:muscles-hand}
\end{subfigure}
\caption{Muscles of the human hand (from~\cite{Young2013Anatomy}, CC BY 4.0).}\label{fig:muscles}
\end{figure}

The human hand has a high force-to-weight ratio.
Grip forces up to 662 N have been measured in 25--29 year old men (dominant hand, 90th percentile)~\cite{Wang2018HandGrip} with a dynamometer, i.e., with all fingers working together to maximize force.

The force produced by muscles is applied to bones through tendons. Muscles interact and work together in complex ways. For instance, flexors and extensors together control finger position, force, and stiffness of the hand~\cite{Controzzi2014Design}.

More than 30 muscles move the hand~\cite{Kapandji2007Physiology} (see Figs.~\ref{fig:muscles}a and \ref{fig:muscles}b). Appendix~\ref{app:human_hand_muscles} lists all muscles and their functions.
The largest muscles that move the hand are located outside of it. These strong, extrinsic muscles (Fig.~\ref{fig:muscles-forearm}) move the wrist joint, fingers 2--5 simultaneously, and fingers 1/2/5 individually.
Looking at the wrist joint muscles, we observe that multiple muscles are involved per movement\footnote{\textit{Flexor carpi radialis}, \textit{flexor carpi ulnaris}, and \textit{palmaris longus} flex the wrist joint. \textit{Extensor carpi ulnaris}, \textit{extensor carpi radialis longus}, and \textit{extensor carpi radialis brevis} extend the wrist joint \citep{Young2013Anatomy}.} and each muscle has multiple functions.\footnote{All \textit{carpi radialis} muscles contribute to abduction, and all \textit{carpi ulnaris} muscles contribute to adduction of the wrist. Furthermore, there are extrinsic \textit{finger} muscles that contribute to wrist flexion and extension \citep{Young2013Anatomy}.}
The largest extrinsic flexors and extensors of the fingers actuate multiple joints:
flexor digitorum superficialis (FDS),\footnote{Its four tendons attach to the middle phalanges 2--5 and flex fingers and wrist.}
flexor digitorum profundus (FDP),\footnote{Its four tendons attach to the distal phalanges 2--5, passing through the tendons of FDS, and flex fingers and wrist.}
and extensor digitorum communis (EDC).\footnote{Its four tendons attach to the distal phalanges 2--5 and extend fingers the wrist.}

Intrinsic muscles (Fig.~\ref{fig:muscles-hand}) are inside of the human hand.
Interossei muscles (four dorsal, three palmar) abduct/adduct fingers 2--5 and, together with the lumbricals,\footnote{Lumbricals are located between tendons of the FDP at the palm and the back side of the distal phalanges 2--5.} allow individual flexion of the of MCP joints, and extend the PIP and DIP joints.

The thumb has a large range of motion in the TMC joint and ability to oppose the other fingers. The two DoF of the TMC joint are controlled by all eight intrinsic and extrinsic muscles of the thumb.

The little finger has individual extensor, flexor, abductor, and a muscle that brings the little finger into opposition, hence, its abduction can be stronger than that of fingers 2--4, and it can be in opposition, however, less than the thumb.
Furthermore, the palm can be hollowed.

Often multiple non-redundant muscles work together to perform a movement, e.g., producing specific forces with the index finger often requires cocontraction: less than 5\% of feasible forces are robust to loss of any one muscle~\cite{Kutch2011Muscle}.

\subsubsection{Individual Control of Fingers and Joints}
In comparison to monkeys, human fingers move more independently~\cite{Hager-Ross2000Quantifying}. The thumb~\cite{Ingram2008Statistics} and index fingers are the most independently moving fingers, while the ring and middle fingers are most coupled with other fingers~\cite{Hager-Ross2000Quantifying}.

The greater independence of human fingers arises from anatomical differences such as splitting of a separate muscle belly and tendons, and a loss of tendons in multitendoned muscles \cite{Hager-Ross2000Quantifying}, but also from finer representation in the primary motor cortex.

Nevertheless, humans cannot completely isolate movements of each DoF.
Flexion of one finger may cause nearby fingers to move (enslaving force phenomenon \citep{Slobounov2002Role}). The effect increases with the force applied. Mechanical coupling between tendons and muscle compartments (e.g., FDP flexes all three finger joints \citep{Appell2008Funktionelle}), co-activating multi-digit motor units, and overlapping representations in the brain are possible causes.

Angular motion mostly happens in the middle joint of each finger (usually PIP joint, MCP for the thumb)~\cite{Hager-Ross2000Quantifying}.
 Precision grasps of the human hand with small contact areas rely on independent abduction/adduction of the fingers more than on independent flexion/extension, as this allows for precise opposition of the thumb~\cite{Montagnani2016Independent}.

\subsubsection{Soft Cover (Muscles and Skin)}
The soft and deformable cover of the human hand consists mainly of its intrinsic palmar muscles and its skin. By contracting muscles in the palm or flexing fingers, the surface shape can be changed.

The benefit of flexible and soft hands is twofold: (1) robustnesst to contact with the environment, and (2) support of prehension by adaptation to the shape of objects.
Humans actively use and exploit environmental constraints. For example, when grasping a flat object from a table, they often slide it to the edge to grasp it more easily~\cite{DellaSantina2017Postural}.

\subsubsection{The Human Arm}
Our hand is attached to an arm that possesses seven DoF: three in the shoulder, one in the elbow, and three in the wrist. Hence, we have a redundant DoF (see Section \ref{sec:robotic_hands_mechanics}), which allows us to incorporate secondary objectives besides pose control, e.g., avoiding obstacles or torque limits \cite{Elias2024Redundancy,Wang2025Analytical}.

Montagnani et al.\ \cite{Montagnani2015it} show that wrist flexibility is more important than finger dexterity in prosthetics.
Lack of wrist flexibility is compensated by arm and trunk movements~\cite{Major2014Comparison}.

\subsection{Perception and Sensors}
\label{sec:human_perception}
The human hand integrates tactile, proprioceptive, and visual information across skin, muscle, and joint receptors, creating a multimodal sensory representation of both object properties and limb state. This dense sensor network enables the perception of detailed contact geometry and nuanced positional feedback, which in turn supports stable force control and flexible exploration strategies during manipulation \cite{Dahiya2010Tactile,Lederman2009Haptic}. Precise measurement of contact forces is essential for regulating grasp forces and maintaining stable grasps \cite{Dahiya2010Tactile}, and when proprioceptive input is compromised, reach trajectories become unstable, postural stability is impaired, and hand coordination remains poor even in the presence of visual feedback~\cite{Lederman2009Haptic}.

\subsubsection{Sensory Receptors in the Hand}
\begin{figure}[tb]
\centering
\includegraphics[width=0.5\textwidth]{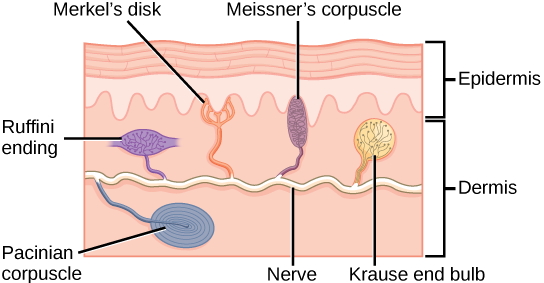}
\caption{Primary mechanoreceptors in the human skin. Merkel's cells respond to light touch, Meissner's corpuscles respond to touch and low-frequency vibrations. Rufinni endings respond to deformations and warmth. Pacinian corpuscles respond to transient pressure and high-frequency vibrations. Krause end bulbs respond to cold. Image from \cite{Clark2020Biology} CC BY 4.0.}
\label{fig:skin}
\end{figure}

Haptic perception is often linked to the sense of touch, mediated primarily by skin receptors that fall into three categories based on their function: mechanoreceptors for pressure and vibration, thermoceptors for temperature changes, and nociceptors for pain \citep[Chap.\ 9]{Purves2012Neuroscience}. These receptors allow us to perceive size, shape, texture, and temperature, which are crucial for object manipulation~\citep{Dahiya2010Tactile}.

The classification of mechanoreceptors in the human hand (illustrated in Fig.~\ref{fig:skin}) includes four primary types, each with distinct characteristics~\cite{Dahiya2010Tactile}.
Pacinian corpuscles (FA II) are fast-adapting receptors that detect high-frequency vibrations, which are important for tool use, with a typical stimulus frequency range of 40--500+ Hz. Ruffini corpuscles (SA II) adapt slowly and respond to skin stretch, providing feedback on finger position, stable grasp, and tangential forces, with effective stimuli frequencies around 100--500+ Hz. Merkel cells (SA I), also known as slow-adapting, have high spatial acuity (approximately 0.5 mm) and are crucial for texture perception and form detection, responding to low-frequency stimuli between 0.4 and 3 Hz. Meissner's corpuscles (FA I) are fast-adapting, sensitive to low-frequency vibrations (3--40 Hz), and are involved in motion detection and grip control, with spatial acuity ranging from 3 to 4 mm. All mechanoreceptors exhibit similar conduction velocities of approximately 35--70 m/s, facilitating the rapid relay of tactile information essential for fine motor tasks. They contribute to detecting fingertip forces, vibrations, and tangential loads \citep{Liu2020Bioinspired}.

Proprioception, or the sense of self-movement and body position, is integral to haptic perception. This sense originates from receptors in muscles, joints, and tendons \citep{Lederman2009Haptic, Dahiya2013Tactile}, playing a key role in perceiving object properties, such as shape, through the alignment of bones and muscle stretching \citep{Navarro-Guerrero2023VisuoHaptic}. Proprioceptive receptors in muscles, skin, and joint capsules \citep{Tabot2013Proprioception} are crucially connected with motor control \citep{Taylor2013Kinesthetic}, and have a fast conduction velocity of 80--120 m/s \citep{Siegel2006Essential}. These proprioceptive signals, interacting with tactile feedback, convey information about the state of hand, wrist, and forearm muscles.

Muscle spindles sense changes in muscle length, while Golgi tendon organs (GTOs) measure muscle-applied force \citep{Tabot2013Proprioception}, contributing to detailed representations of hand and limb positions. Joint capsule receptors measure extreme positions of the joints to prevent overextension \citep{Tabot2013Proprioception}.

Cutaneous nerve fibers on the dorsal hand surface react to skin stretch during joint movements and convey joint angle information \citep{Lutz2021Proprioceptive}. Stimulating the skin on the dorsal surface can induce illusions of movement. However, this effect has been tested only when a single joint was deflected \citep{Lutz2021Proprioceptive}.

\subsubsection{Signal Integration}
The brain devotes a significant portion of the somatosensory cortex to the hand: about 20\% of the surface in Brodmann's areas~\cite{Lutz2021Proprioceptive} are allocated to hand representations. Stereognosis, the haptic perception of three-dimensional shapes, requires integrating proprioceptive with cutaneous signals~\cite{Lutz2021Proprioceptive}. Proprioceptive signals describe hand conformation during grasp, while tactile signals deliver contact location and geometric features such as edges and corners at each contact point~\cite{Lutz2021Proprioceptive}. This integration ensures effective object perception and showcases how multimodal signals encode object identity even before contact occurs~\cite{Lutz2021Proprioceptive}.

\subsubsection{Role of the Modalities}
\paragraph{Visual feedback}
Dexterous manipulation relies on the coordinated use of visual, tactile, and proprioceptive information, which together provide rich estimates of object properties and limb state during interaction \citep{Land2000Eye,Lederman2009Haptic,Dahiya2010Tactile}. Visual feedback supplies global information about the external scene and object pose and is particularly important when coordinating hand movements with moving or distant objects \citep{Land2000Eye}. However, when vision is impaired or occluded, fine-manipulation performance degrades, and an exclusive reliance on visual feedback increases cognitive load and processing time, because visual information must be interpreted and transformed into motor commands \citep{Johansson1992Somatosensory,Odoh2024Performance}. In robotic systems, high-quality visual processing is also computationally demanding, which motivates distributing perceptual load across dedicated tactile and proprioceptive sensors \citep{Goyal2023RVT}.

\paragraph{Haptic feedback}
Proprioceptive and tactile channels provide complementary information that is essential for fluent manipulation \citep{Lederman2009Haptic}. Proprioception supports coordinated multijoint movements, precise timing, and motor learning, and when it is impaired reaches become inaccurate and trajectories unstable even with intact vision \citep{Ghez1995Impairments,Rossi2021Mechanisms}. Tactile feedback enables real-time modulation of grip force, slip prevention, and texture and material discrimination, complementing visual and proprioceptive feedback in dynamic environments \citep{Johansson2009Coding,Lederman2009Haptic}.

\subsubsection{Role in Specific Manipulation Tasks}
We examine more closely the role of sensors in two specific manipulation tasks: object exploration and grasping.

\paragraph{Object exploration}
\begin{figure}[tb]
\centering
\includegraphics[height=0.12\textheight]{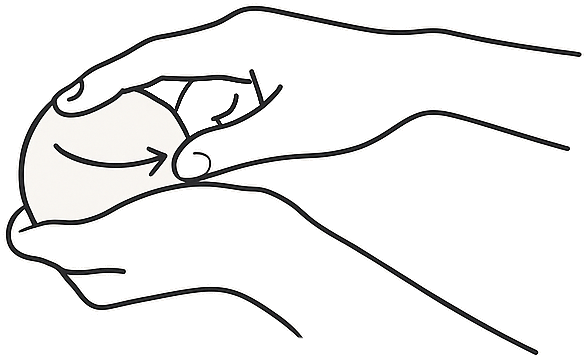}
\includegraphics[height=0.12\textheight]{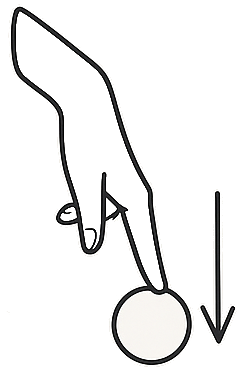}\\
\includegraphics[height=0.12\textheight]{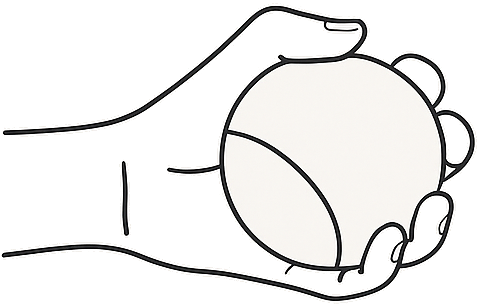}
\includegraphics[height=0.12\textheight]{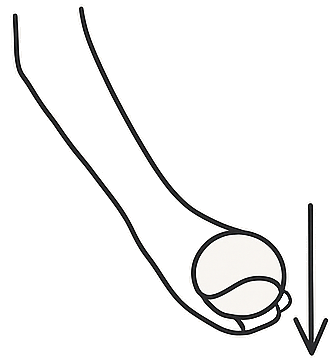}\\
\includegraphics[height=0.12\textheight]{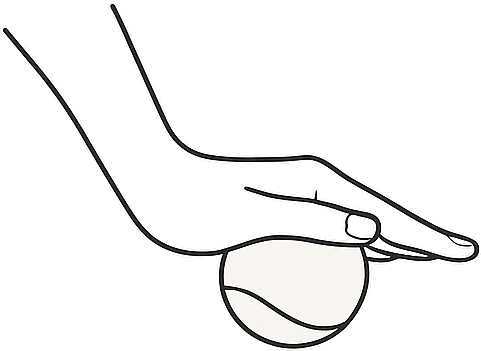}
\includegraphics[height=0.12\textheight]{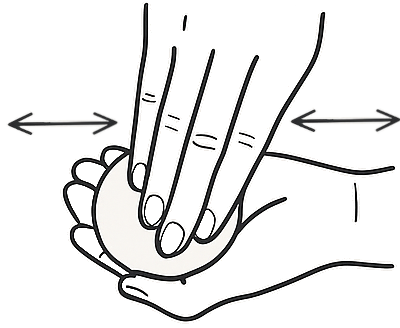}
\caption{Illustration of six exploratory procedures~\cite{Lederman2009Haptic}. From left to right and top to bottom: Contour Following, Pressure, Enclosure, Unsupported Holding, Static Contact, and Lateral Motion. Adapted from \cite{Nelinger2015Tactile}, CC BY 3.0.}
\label{fig:exploratory-procedures}
\end{figure}
Haptic perception enables humans to explore objects using strategies known as exploratory procedures (EPs)~\cite{Lederman1987Hand,Lederman2009Haptic}. EPs are categorized based on their focus: the substance of an object (texture, hardness, temperature, weight), structural properties (global shape, volume, weight), and function discovery (movable parts, potential functionality). Fig.~\ref{fig:exploratory-procedures} shows examples of EPs from the first two categories. The lateral motion EP moves skin over an object's surface to assess texture, while the pressure EP tests hardness through poking or tapping. Static contact EP assesses temperature. Other EPs include unsupported holding for weight inference, enclosing for global shape, contour-following for detail, and part-motion testing and function testing to understand movement and function.

\paragraph{Grasping}
Cutaneous receptors are central to reaching and grasping: they determine shear and load forces to regulate grip force, and their afferent responses mark transitions between the reach, load, lift, and hold phases so that planning and control in the brain are organized around mechanical events~\citep{Dahiya2010Tactile}. Without tactile feedback, internal models for anticipatory control become outdated, the finger-opening phase is prolonged, and grasp timing degrades. Force direction matters as much as magnitude: it balances normal against tangential components (the ``friction cone'') and conveys shape, surface texture, slip, and material properties \citep{Dahiya2010Tactile}.

\section{Robotic Hands}
Having characterized the human hand in terms of biomechanics and perception, we compare current robotic and prosthetic hands along the same dimensions: mechanical design and sensory capabilities. We use \textit{gripper} for two-fingered devices with few actuators and \textit{robotic hand} for multi-finger designs with multiple DoF and actuators. This review focuses on rigid and semi-rigid hands, which dominate both robotics research and prosthetics. A comprehensive overview of individual hands is provided in the supplementary material. The gap between human and robotic capabilities motivates the research questions in the next section.

\begin{figure*}[tb]
\centering
\includegraphics[width=0.6\textwidth]{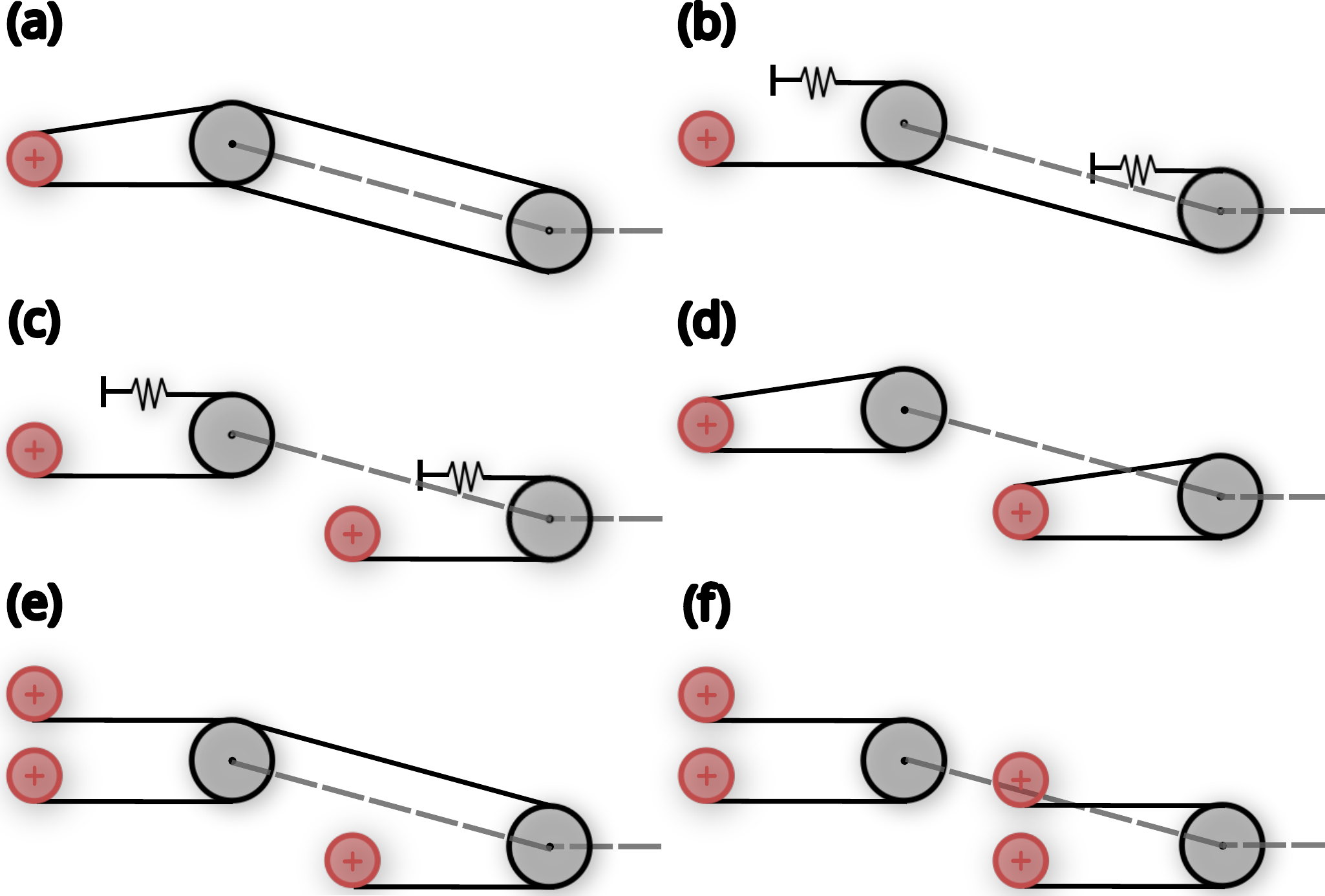}
\caption{Representation of possible robotic hand kinematic architectures (adapted from Controzzi et al.~\cite{Controzzi2014Design}). Notation: $M$---number of actuated tendons (red circles); $N$---number of bidirectional DoFs (black circles). Diagrams show: (a) $M=N$: coupled joints; (b) $M<N$: underactuated transmission; (c) $M=N$: fully actuated open chain; (d) $M=N$: fully actuated closed chain; (e) $M=N+1$: fully controllable; (f) $M=2N$: agonist/antagonist transmission \cite{Controzzi2014Design}. Continuous lines indicate tendons, dashed lines joints, and zigzag lines springs.}
\label{fig:kinematics_controzzi}
\end{figure*}

\subsection{Robotic Hand Mechanics}
\label{sec:robotic_hands_mechanics}
Robotic hands are designed in various ways depending on their intended use. Controzzi et al.~\cite{Controzzi2014Design} suggest considering six aspects when designing a robotic hand: (1) hand kinematics, (2) actuation principle, (3) actuation transmission, (4) sensors, (5) materials, and (6) manufacturing method.

\subsubsection{Fingers}
A central challenge in hand design is the optimal number of fingers required for stable grasping versus in-hand manipulation. While five fingers (as seen in the \textit{Shadow Dexterous Hand}) maximize anthropomorphism, they introduce significant control complexity and mechanical packaging challenges. Conversely, three-fingered hands (like the \textit{BarrettHand}) are frequently employed to achieve stable tripod grasps, offering a significant dexterity improvement over simple parallel-jaw grippers.
However, non-anthropomorphic architectures with four fingers were also considered: designs such as the Yale OpenHand~\cite{Ma2014underactuated} or the Tesollo DG-4F hand~\cite{TesolloDG4F} introduce the mechanical redundancy necessary for \textit{finger gaiting}, by using two opposing finger pairs. This configuration reduces the actuator count compared to a five-fingered hand, yet provides significantly higher manipulability and grasp safety than a three-fingered gripper, particularly during object reorientation tasks. The non-anthropomorphic concept of two opposing fingers mitigates obstructions and workspace reduction during in-hand manipulation.

\subsubsection{Joints and Degrees of Freedom}
The required DoFs and the number of actuators determine the hand's kinematics~\citep{Controzzi2014Design}. The mechanical fingers/grippers are mainly composed of different joints connected in a serial manner, allowing them to perform a specific task. In order to be able to control $n$ DoFs, we need $m= n + 1$ independent actuation tendons~\citep{Controzzi2014Design}. Such kinematic architecture is used in the \textit{BarrettHand}.\footnote{\url{https://advanced.barrett.com/barretthand}}
Nevertheless, hands with actuation tendons that exceed $n + 1$ are more versatile and dexterous. These kinematic architectures are called redundant and are used, for example, in \textit{The Shadow Dexterous Hand}\footnote{\url{https://www.shadowrobot.com}}.
Regardless of the advantages of using redundant transmission, it is also possible to design hands with underactuation ($m < n$) and coupled transmission. Such a design is inspired by the human hand, as shown in Section~\ref{sec:humanHand_biomechanics}. For instance, recent advancements in prosthetic design have explored six-bar linkage mechanisms to create bioinspired, kinematically coupled 1-DoF fingers that mimic natural human flexion~\cite{Ceccarelli2025Design,Ceccarelli2025Atlas}.

Closely related to the concept of underactuation is the development of semi-rigid, or compliant-joint, robotic hands. Instead of traditional revolute joints, these designs utilize compliant flexure mechanisms to connect rigid structural links. This hybrid architecture allows the fingers to achieve passive adaptability, naturally conforming to complex object geometries upon contact. As a result, semi-rigid hands provide safe and flexible grasping without requiring fully soft elastomer bodies or highly complex control algorithms, offering a practical balance between traditional structural payload capacity and the compliance of soft robotics~\cite{Birglen2008Underactuated}. Prominent examples of this approach include the Yale OpenHand project~\cite{Ma2014underactuated}, which employs elastomeric joints connecting rigid finger segments to dynamically adjust to target objects.

To systematically categorize these diverse design choices, spanning from fully actuated and redundant systems to underactuated and compliant mechanisms, Controzzi et al.~\cite{Controzzi2014Design} provide six possible kinematic architectures for robotic hands, depicted in Fig.~\ref{fig:kinematics_controzzi}.

\begin{figure}
\centering
\begin{subfigure}[t]{\columnwidth}
\centering
\includegraphics[width=\textwidth]{Fig7a}
\caption{Legend for kinematic diagrams.}
\label{fig:kin_legend}
\end{subfigure}
\begin{subfigure}[t]{\columnwidth}
\centering
\includegraphics[scale=0.6]{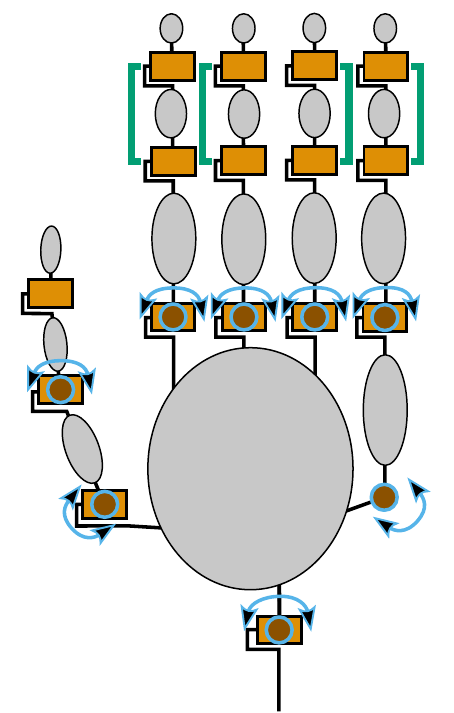}
\caption{Shadow Dexterous Hand.}
\label{fig:kin_shadow_dexterous_hand}
\end{subfigure}
\begin{subfigure}[t]{\columnwidth}
\centering
\includegraphics[scale=0.7]{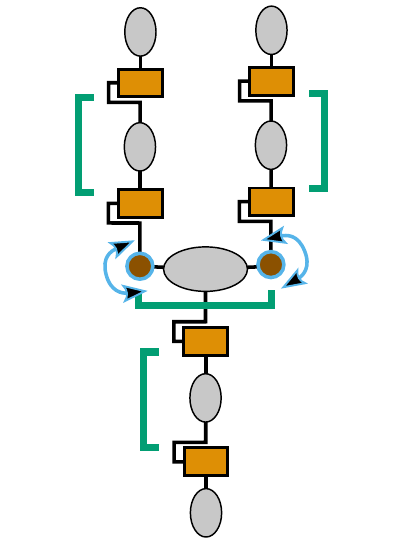}
\caption{BarrettHand.}
\label{fig:kin_barretthand}
\end{subfigure}

\caption{Kinematic diagrams of representative robotic hands. Notation based on Wang et al.~\citep{Wang2008Topology}.}
\label{fig:kin_diagrams}
\end{figure}

\textit{The Shadow Dexterous Hand} is one of the most advanced robotic hands and is produced by \textit{Shadow Robot}. It has five fingers and 24 DoFs, of which 20 are controllable. In addition, similar to the human hand, it is possible to perform opposition and reposition of the thumb, flexion, extension, adduction, and abduction of all fingers. Such movements enable a dexterous hand that can perform various complex manipulation tasks.
Fig.~\ref{fig:kin_shadow_dexterous_hand} represents the kinematic architecture of the Shadow Dexterous Hand using the kinematic notations derived from~\cite{Wang2008Topology} and depicted in Fig.~\ref{fig:kin_legend}.

A less complex robotic hand is the \textit{BarrettHand}. It has three fingers and eight DoFs, although only four of these are independently actuated using four motors. All fingers support flexion and extension movements. Opposition and reposition are possible with two fingers. The kinematic architecture of the BarrettHand is depicted in Fig.~\ref{fig:kin_barretthand}.

\subsubsection{Actuators and Transmission}
Apart from the chosen kinematic architecture, the transmission actuator type and mechanism do impact the performance of the robotic hand. For instance, actuators used in the artificial hand can be electrical (e.g., DC motors), pneumatic, or hydraulic. Natural muscles have a power density ($\rho$) of approximately $500 W/Kg$. Such power density can be achieved by hydraulic or pneumatic actuators (respectively, $2000 W/Kg$ and $400 W/Kg$)~\citep{Huber1997Selection}. Such solutions are already used in artificial robotic hands (e.g., GRIPKIT Industrial (P PRO)\footnote{\url{https://weiss-robotics.com/gripkit-for-cobots/}}, EHA hand~\citep{Ko2017Underactuated}). However, they are usually bulky or hard to maintain. DC motors (electrical actuators), on the other hand, can reach a power density $\rho = 100 W/Kg$~\citep{Huber1997Selection}. While these actuation mechanisms may have a lower power-to-weight ratio compared to others, they remain popular actuators thanks to their compact size and cost-effective maintenance~\citep{Controzzi2014Design}.

In order to accurately control joints, transmission is crucial. Such systems enable converting the power provided by the actuators to a specific hand/finger movement. An ideal transmission system would have low inertia, friction and backlash, and would be overall compact and have low weight. Several transmission types can be deployed: including tendons, gear trains, belts, linkages or flexible shafts~\cite{Controzzi2014Design}. In the human hand, we found tendons that connect muscles to bones~\cite{Tanrikulu2015Anatomy}. Such tendons run in sheaths. A similar system is commonly found in artificial robotic hands, as such a design reduces friction and enables remote control of the fingers (e.g., Shadow Dexterous Hand). For further details on transmission, we refer to Controzzi et al.~\cite{Controzzi2014Design}.

\subsubsection{Robotic Arms}
Furthermore, robotic hands are attached to robotic arms. Six joints are sufficient to control all position and orientation components in 3D space so that more DoF are redundant. However, a seventh, redundant DoF allows for avoiding singularities, obstacles, joint limits, or torque limits \cite{Elias2024Redundancy,Wang2025Analytical}. Hence, arm design should not be neglected when developing dexterous manipulation systems.

\subsubsection{Review of Robotic Hands}

\begin{figure}[tbp]
\centering
\begin{subfigure}[t]{\columnwidth}
\centering
\scalebox{0.32}{\includegraphics[clip=true,trim=0 0 0 5]{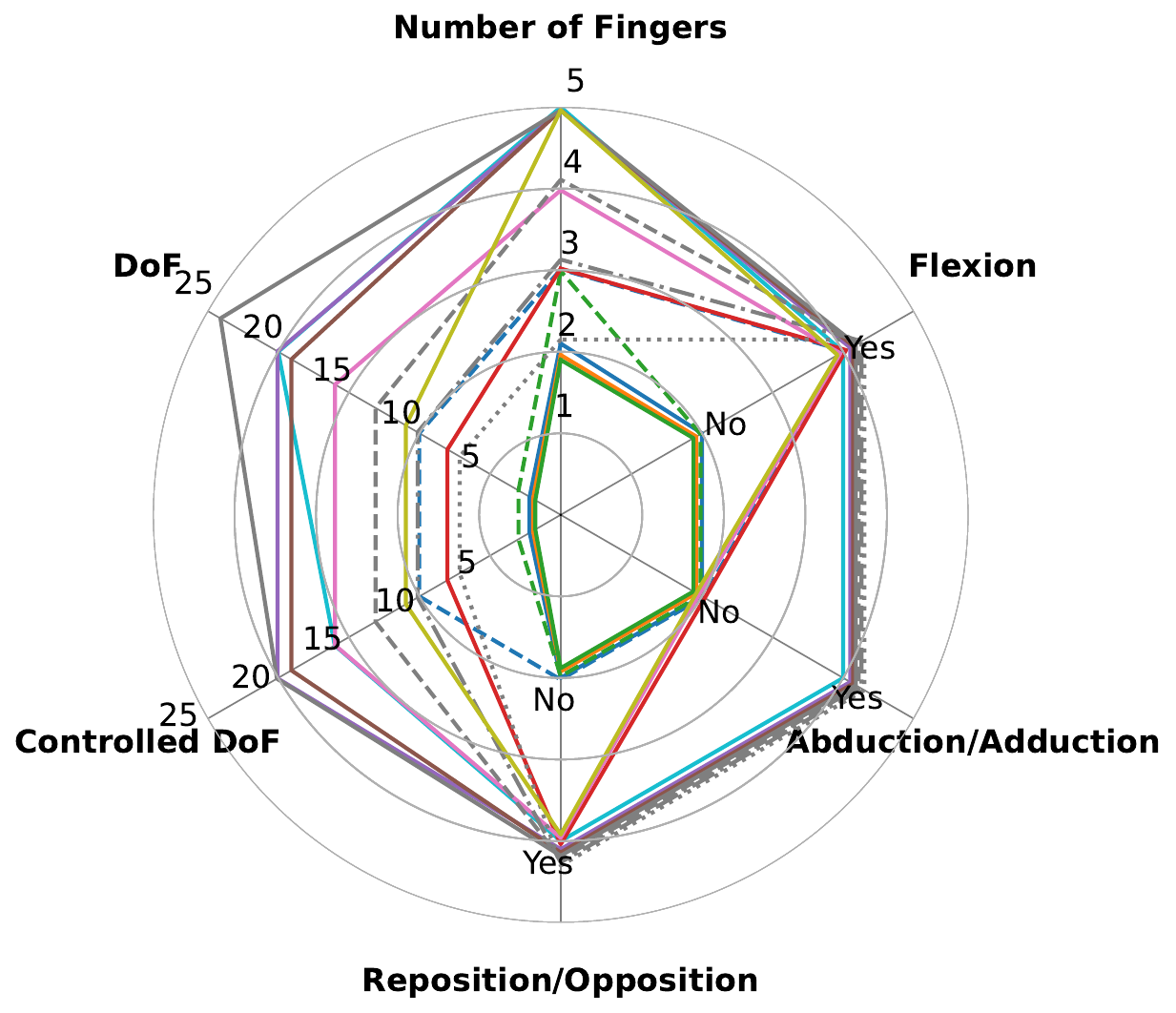}}
\scalebox{0.28}{\includegraphics{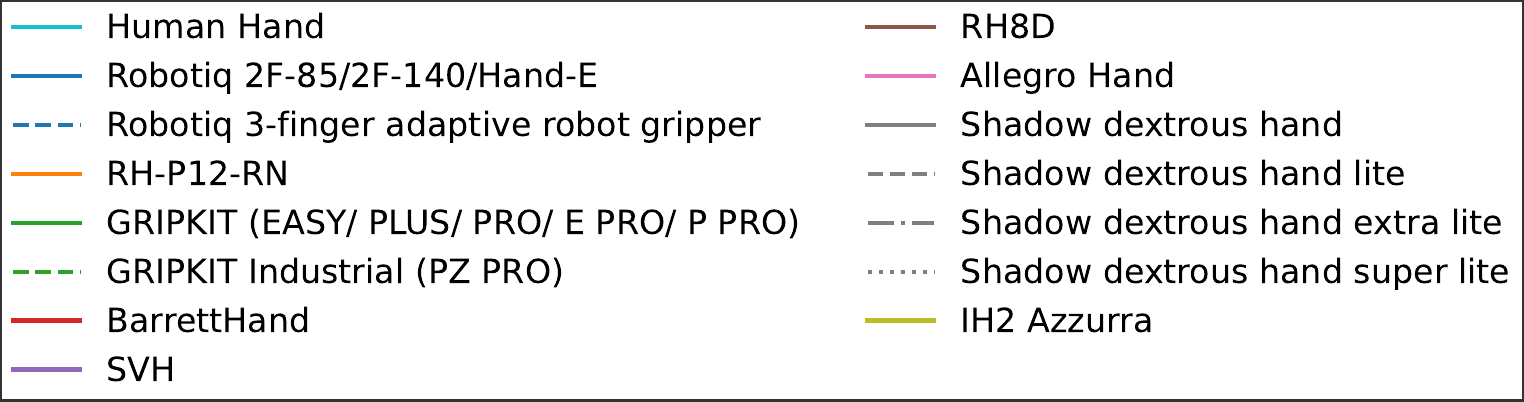}}
\caption{Radar diagram of selected robotic hands. The same color is used for hands from the same manufacturer.}
\label{fig:non-pros-hands_radar}
\end{subfigure}
\begin{subfigure}[t]{\columnwidth}
\centering
\scalebox{0.32}{\includegraphics[clip=true,trim=0 0 0 5]{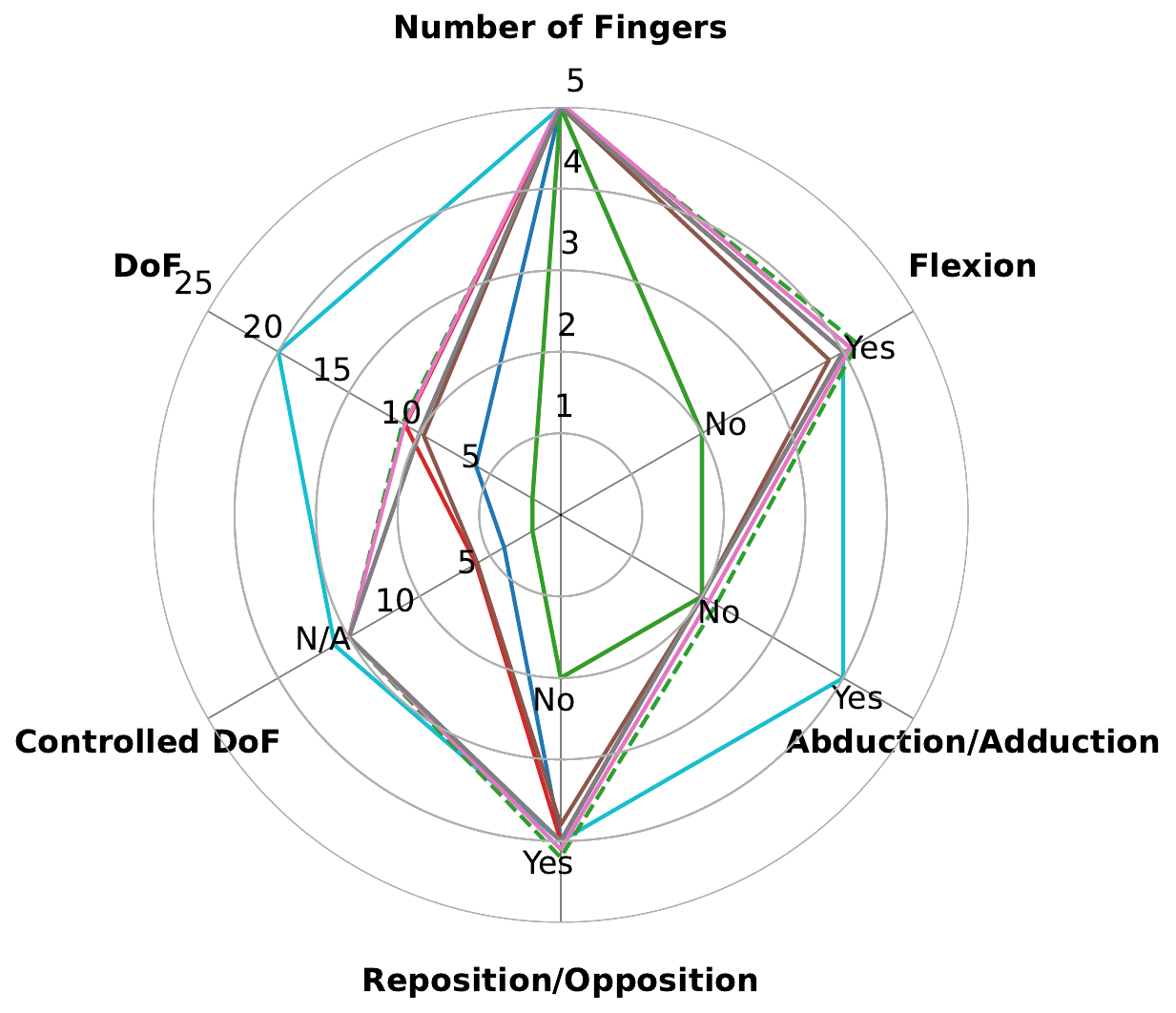}}
\scalebox{0.28}{\includegraphics{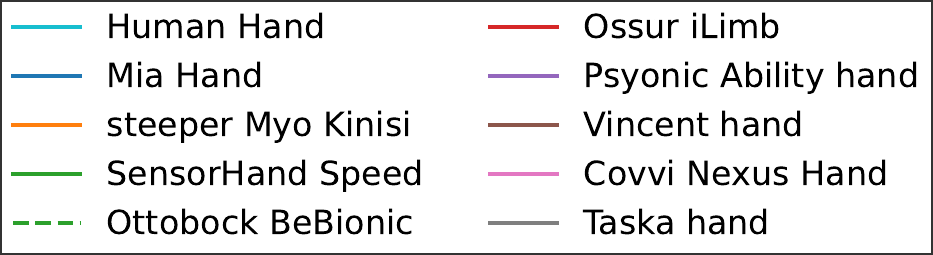}}
\caption{Radar diagram of some selected prosthetic hands.}
\label{fig:pros-hands_radar}
\end{subfigure}
\caption{Radar diagrams of prosthetic and robotic hands. }
\label{fig:hand_radar}
\end{figure}

We reviewed various robotic hands used in industry, research, and also as prosthetic hands. These devices were selected based on their relevance, popularity in current robotics research, and commercial availability at the time of writing. Our goal was to compile a representative sample of the most widely used prototypes rather than an exhaustive market survey. We first categorize these into robotic and prosthetic hands based on the primary application for which the manufacturer/researcher intended the hand to be designed.

We focus on six different features: DoF, number of actuators, number of fingers, possibility of flexion/extension, adduction/abduction, and opposition/reposition.
For a detailed description of each robotic hand, please refer to the supplementary material.
The characteristics of robotic and prosthetic hands are visualized in Fig.~\ref{fig:hand_radar}.

To synthesize the design choices discussed above, Table~\ref{tab:design_comparison} summarizes the key trade-offs between different kinematic and actuation architectures found in current state-of-the-art robotic and prosthetic hands.

From a mechanical perspective, the most advanced robotic hands, such as the Shadow Dexterous Hand, can replicate the number of fingers and DoF found in a human hand. However, they lack the flexibility, deformability, and the passive range of motion inherent in human joints. Achieving high dexterity often comes at the expense of reduced force production. This trade-off between dexterity and force limits the overall performance of robotic hands, highlighting the need for further research in areas such as material science, mechanical design, and control systems.

\begin{table*}[btp]
	\footnotesize
	\centering
	\renewcommand{\arraystretch}{1.2}
	\setlength{\tabcolsep}{5pt}
	\caption{Comparison of common robotic hand design architectures: Advantages and Disadvantages.}
  \label{tab:design_comparison}
  \begin{tabular}{@{}p{0.20\textwidth} p{0.34\textwidth} p{0.41\textwidth}@{}}
  \toprule
  \textbf{Design Architecture} & \textbf{Advantages} & \textbf{Disadvantages} \\ \midrule \midrule
  \multicolumn{3}{@{}l}{\textit{\textbf{Actuation Principle}}} \\ \midrule
  Fully Actuated\newline(e.g., Barrett Hand)
  & \vspace{-0.4\topsep}
  \begin{itemize}[leftmargin=*,noitemsep,nolistsep]
  \item High dexterity and manipulability~\cite{Controzzi2014Design}
  \item Independent control of each joint~\cite{Piazza2019Century}
  \item Capable of complex in-hand manipulation~\cite{Morgan2022Complex}
  \end{itemize}
  & \vspace{-0.4\topsep}
  \begin{itemize}[leftmargin=*,noitemsep,nolistsep]
  \item High control complexity (high dimensionality)~\cite{Controzzi2014Design}
  \item Increased weight, cost, and size~\cite{Huber1997Selection}
  \item Higher energy consumption~\cite{Piazza2019Century}
  \end{itemize}
  \\ \midrule
  Underactuated\newline(e.g., Prosthetic)
  & \vspace{-0.4\topsep}
  \begin{itemize}[leftmargin=*,noitemsep,nolistsep]
  \item Mechanical simplicity and lower weight~\cite{Birglen2008Underactuated}
  \item Passive adaptability to object shape~\cite{Ko2017Underactuated, Odhner2014Compliant}
  \item Lower cost and easier control~\cite{Birglen2008Underactuated}
  \end{itemize}
  & \vspace{-0.4\topsep}
  \begin{itemize}[leftmargin=*,noitemsep,nolistsep]
  \item Reduced dexterity (cannot control all joints)~\cite{Montagnani2016Independent}
  \item Difficulty performing precise in-hand manipulation~\cite{Ma2017Yale}
  \item Predictability of grasp can be lower~\cite{Birglen2008Underactuated}
  \end{itemize}
  \\ \midrule
  \multicolumn{3}{@{}l}{\textit{\textbf{Transmission Type}}} \\ \midrule
  Tendon-Driven
  & \vspace{-0.4\topsep}
  \begin{itemize}[leftmargin=*,noitemsep,nolistsep]
  \item Motors can be placed remotely (forearm) reducing hand inertia~\cite{Controzzi2014Design}
  \item Mimics biological systems~\cite{Piazza2019Century}
  \item Allows for slim finger design~\cite{Controzzi2014Design}
  \end{itemize}
  & \vspace{-0.4\topsep}
  \begin{itemize}[leftmargin=*,noitemsep,nolistsep]
  \item Issues with friction, elasticity, and hysteresis~\cite{Piazza2019Century}
  \item Complex maintenance (tendon routing/breakage)~\cite{Controzzi2014Design}
  \item Requires tensioning mechanisms~\cite{Controzzi2014Design}
  \end{itemize}
  \\ \midrule
  Linkage/Gear-Driven
  & \vspace{-0.4\topsep}
  \begin{itemize}[leftmargin=*,noitemsep,nolistsep]
  \item High structural stiffness and precision~\cite{Piazza2019Century}
  \item Robust force transmission~\cite{Controzzi2014Design}
  \item Less maintenance required than tendons~\cite{Piazza2019Century}
  \end{itemize}
  & \vspace{-0.4\topsep}
  \begin{itemize}[leftmargin=*,noitemsep,nolistsep]
  \item Higher backlash (unless direct drive)~\cite{Controzzi2014Design}
  \item Actuators usually located in the hand, increasing inertia~\cite{Piazza2019Century}
  \item Rigid collisions can damage gears~\cite{Piazza2019Century}
  \end{itemize}
  \\ \bottomrule
  \end{tabular}
\end{table*}

\subsubsection{Current Trends}
Beyond the established categories and traditional kinematic designs, the state of the art is currently shifting toward three major hardware development trends driven by computational and mechanical necessities:

(1) AI-Driven Hardware Design: The integration of Artificial Intelligence (AI) and Reinforcement Learning (RL) is heavily influencing hardware specifications. Modern hands are increasingly designed to be mechanically robust to withstand the high-cycle, trial-and-error collisions inherent to learning algorithms \cite{OpenAI2019Solving}. Furthermore, there is a growing focus on developing the physical hardware alongside the software, ensuring that the mechanical design itself simplifies the learning process~\cite{Fay2025House,Yang2024Evolving}.

(2) Direct and Quasi-Direct Drive Actuation: To prioritize bandwidth and back-drivability, recent designs utilize high-torque motors with either no gearing (Direct Drive) or low gear ratios (Quasi-Direct Drive, typically $<10:1$). This architecture provides \textit{impact transparency}, allowing for force estimation without expensive torque sensors. Notable examples include the CMU Direct Drive Hand \cite{Bhatia2019Direct}, which uses a 1:1 transmission for maximum transparency, and the Blue grippers \cite{Gealy2019QuasiDirect}, which utilize QDD to balance torque and compliance.

(3) Soft Robotics: There is a distinct move toward compliant materials and fluidic or pneumatic actuation. By utilizing elastomers and variable stiffness mechanisms, these hands ensure safe human-robot interaction and passive adaptability to object geometries, reducing control complexity for delicate grasping tasks \cite{Rus2015Design, Yasa2023Overview}. However, as previously established, purely soft robotic end-effectors fall outside the primary scope of this paper, which remains focused on examining traditional rigid and semi-rigid kinematic architectures.

\subsection{Perception and Sensors}
\label{sec:robot_perception}
To perform dexterous manipulation, a robot must understand what the object being manipulated is and the type of operation required, that is, the task requirements \citep{Xia2022Review}. This understanding depends on sensing modalities that provide information about contact events, object properties, and the manipulator’s state.

\subsubsection{Sensory Receptors}
Tactile sensors are primarily designed to mimic mechanoreceptors, particularly for detecting mechanical pressure. The main objectives of tactile sensors are to determine the location, shape, and intensity of contacts. These properties are determined by measuring the instantaneous pressure or force applied to the sensor's surface at multiple contact points. In addition to quasi-static contact information, the late effects of contact, that is, body-borne vibrations, may carry relevant information. Body-borne vibrations are not as commonly measured or exploited as part of haptic sensing, but there are examples, and this is becoming an active area of research \citep[e.g.,][]{ZaiElAmri2026VibroSense, JuinaQuilachamin2023Biomimetic, Bonner2021AU, Toprak2018Evaluating}. This includes sensors inspired by hair follicle receptors or ciliary structures \citep{Alfadhel2015Magnetic, Kamat2019Bioinspired} that have been shown to be effective in obtaining information about the texture of objects \citep{Ribeiro2020Highly, Ribeiro2020Fruit}.

Thermoceptors are typically not classified as tactile sensors in robotics but are sometimes added to compensate for thermal drift~\citep{Tomo2016Design} or to support material classification through temperature differences~\citep{Wade2017Force,Bhattacharjee2021Material}. Nociceptors have not been developed as dedicated hardware. Instead, nociceptive mechanisms are implemented in software to extend robot lifespan, reduce maintenance, and improve positioning safety and accuracy~\citep{Navarro-Guerrero2017Improving,Navarro-Guerrero2017Effects}. Explicit hardware nociceptors may not be necessary if nociception can be inferred from other sensory modalities.

\subsubsection{Signal Integration}

As in humans (Section~\ref{sec:human_perception}), proprioceptive and tactile sensing provide complementary information streams. Proprioception supports coordination, stability, and motor learning, while tactile feedback enables texture and material discrimination, slip detection, and fine grip-force modulation \citep{Lederman2009Haptic}. For manipulation tasks that hinge on material handling, texture discrimination, or precise grip-force control, tactile feedback is therefore a particularly valuable complement to vision and proprioception.

\subsubsection{Specific Use Cases}
The dexterous robotic hand can achieve more accurate adjustments with a combination of real-time tactile feedback and the control system~e.g. \cite{yuan2024robot,she2021cable}. The sense of touch is no longer simply ``telling'' the robot about contact information and object information. Robot hands use tactile information to make judgments about fingers and control the object's state to reach the target state. In control-level perception, related research topics include tactile servoing, slip detection, and grasp quality measures \citep{Xia2022Review}.

\paragraph{Tactile servoing} In the process of dexterous manipulation, the tactile servo can be assigned in the control architecture, where the motor is driven by tactile feedback. Meanwhile, tactile servoing also provides the robotic hand with a new direction for exploring environments more safely \citep{Xia2022Review}.

\paragraph{Slip detection} Dexterous manipulation with a multi-fingered robotic hand is a challenging task due to the existence of uncertainties arising from sensor noise, slippage, or external disturbances. External disturbances, such as environmental changes, may cause an expected stable grasp to become unstable. Humans can react quickly to instabilities through tactile sensing. The ability to detect slips in tactile perception is critically essential for further dexterous manipulations \citep{Xia2022Review}.

\paragraph{Grasp quality measure} The correct grasp of objects is critically important for dexterous manipulation. Planning a good grasp involves determining grasping points on the object surface and selecting suitable hand configurations.
Given an object and a hand, the key step in grasping planning is to measure the grasp quality \citep{Xia2022Review}.

\begin{table*}[bt]
  \footnotesize
  \centering
  \renewcommand{\arraystretch}{1.2}
  \setlength{\tabcolsep}{4pt}
  \caption{Comparisons of typical tactile sensors technologies. This table is based on Chi et al.\ \cite{Chi2018Recent}, Qu et al.\ \cite{Qu2023Recent}, and Meribout et al.\ \cite{Meribout2024Tactile}.}
  \label{tab:transduction-mechanisms}
  \begin{tabular}[t]{@{}p{0.56\textwidth} p{0.4\textwidth}@{}}
  \toprule
  \textbf{Transduction Mechanisms} & \textbf{Performance Metrics} \\ \midrule \midrule

Capacitive tactile sensors measure changes in capacitance between parallel plates separated by a deformable dielectric, so applied forces modify plate spacing or overlap. They offer high sensitivity and spatial resolution over a large dynamic range with relatively low power consumption. Their main drawbacks are complex readout electronics, susceptibility to stray and electromagnetic noise, cross-talk, and limited durability.
  & \vspace{-0.4\topsep}
  \begin{itemize}[leftmargin=*,noitemsep,nolistsep]
    \item Spatial resolution $\rightarrow$ $15 \mu m$ \cite{Bai2023robotic}
    \item Resolution $\rightarrow$ $3$Pa \cite{Qu2023Recent}
    \item Sensitivity $\rightarrow$ $0.011-6.583$ kPa$^{-1}$ \cite{Qu2023Recent}
    \item Dynamic range $\rightarrow$ $0-1700$kPa \cite{Qu2023Recent}
    \item Response time $\rightarrow$ $48-180$ms \cite{Qu2023Recent}
    \item Hysteresis $\rightarrow$ 2.94\% and creep of 3.11\% \cite{Chen2025Capacitive}
    \end{itemize}
  \\[-0.3\topsep]
  \midrule

Piezoresistive tactile sensors transduce applied forces into  resistance in a deformable conductive or semiconductive element. They offer simple construction, high spatial resolution, low cost, and a high signal-to-noise ratio, and can be fabricated over large or small areas with relatively low latency. Their main drawbacks are hysteresis, relatively high power consumption compared to capacitive and piezoelectric designs, susceptibility to electromagnetic noise, and limited repeatability in manufacturing.
  & \vspace{-0.4\topsep}
    \begin{itemize}[leftmargin=*,noitemsep,nolistsep]
    \item Spatial resolution (4mm \cite{Li2025Design})
    \item Resolution $\rightarrow$ $50$Pa \cite{Qu2023Recent}
    \item Sensitivity $\rightarrow$ $0.0835-133.1$ kPa$^{-1}$ \cite{Qu2023Recent}
    \item Dynamic range $\rightarrow$ $0.8-800$kPa \cite{Qu2023Recent}
    \item Response time $\rightarrow$ $50-90$ms \cite{Qu2023Recent}
    \item Hysteresis $\rightarrow$ 1.05\% \cite{Saxena2024Recent}
    \end{itemize}
  \\[-0.1\topsep]
  \midrule

Piezoelectric tactile sensors generate voltage changes in a piezoelectric material in response to applied mechanical stress. They provide high sensitivity, large dynamic range, high accuracy, and excellent high‑frequency response, making them suitable for detecting rapid transients and vibrations. However, they offer relatively low spatial resolution, suffer from charge leakage, cannot measure static loads, and implementations can be bulky.
  & \vspace{-0.4\topsep}
    \begin{itemize}[leftmargin=*,noitemsep,nolistsep]
    \item Spatial resolution $\rightarrow$ 0.2-2.5mm \cite{Meribout2024Tactile}
    \item Resolution
    \item Sensitivity $\rightarrow$ $0.005$Pa \cite{Qu2023Recent}
    \item Dynamic range $\rightarrow$ $1-750$kPa \cite{Qu2023Recent}
    \item Response time $\rightarrow$ $0.1-60$ms \cite{Qu2023Recent}
    \item Hysteresis
    \end{itemize}
  \\[-0.1\topsep]
  \midrule

Magnetic tactile sensors detect changes in magnetic flux or field caused by deformation of a soft medium containing a magnet or magnetic filler, which are read out by Hall‑effect or giant magnetoresistance sensors. They offer high sensitivity to small normal and shear forces, low hysteresis, and good repeatability, and can provide full 3D force information at fingertip scales. Their main drawbacks are complex fabrication and magnetization processes, limited to low force ranges, and gradual loss of magnetization over time, which requires frequent recalibration.
  & \vspace{-0.4\topsep}
    \begin{itemize}[leftmargin=*,noitemsep,nolistsep]
    \item Spatial resolution $\rightarrow$ 1.2mm \cite{Hu2024Largearea}
    \item Resolution
    \item Sensitivity $\rightarrow$ 3 mN \cite{Meribout2024Tactile}
    \item Dynamic range $\rightarrow$ $0-5$kPa \cite{Meribout2024Tactile}
    \item Response time $\rightarrow$ 65ms \cite{Hu2024Largearea}
    \item Hysteresis
    \end{itemize}
  \\[-0.1\topsep]
  \midrule

Optical tactile sensors based on fiber Bragg gratings encode contact-induced strain as shifts in reflected wavelength or light intensity along an optical fiber. They offer high spatial resolution, wide sensing range, high repeatability, and high sensitivity with excellent immunity to electromagnetic interference, vibration, corrosion, and extreme temperatures, and can be scaled to dense arrays. However, they are typically non-conformable and relatively bulky, and require precise alignment and temperature compensation with complex demodulation electronics.
  & \vspace{-0.4\topsep}
    \begin{itemize}[leftmargin=*,noitemsep,nolistsep]
    \item Spatial resolution $\rightarrow$ 0.77 mm \cite{Lyu2022HighSpatialResolution}
    \item Resolution $\rightarrow$ 0.93mN \cite{Lyu2025Fiber}
    \item Sensitivity $\rightarrow$ 0.01nmN$^{-1}$ \cite{Lyu2025Fiber}
    \item Dynamic range $\rightarrow$ $0-50$N \cite{Lyu2025Fiber}
    \item Response time $\rightarrow$ 10$\mu s$ \cite{Lyu2025Fiber}
    \item Hysteresis
    \end{itemize}
  \\[-0.1\topsep]
  \midrule

Vision-based tactile sensors infer contact forces from deformations of an illuminated compliant membrane or gel, which are imaged by an RGB camera \citep{Yuan2017GelSight,Lambeta2020DIGIT}. They provide very high spatial resolution, high sensitivity, and large dynamic range, since each image pixel effectively acts as a taxel. However, they require an internal light source and camera, are relatively bulky, difficult to deploy on large or highly curved surfaces, and mainly provide indirect tactile information via images rather than direct force measurements.
  & \vspace{-0.4\topsep}
    \begin{itemize}[leftmargin=*,noitemsep,nolistsep]
    \item Spatial resolution $\rightarrow$ $33\mu m$ \cite{Xin2025Visionbased}
    \item Resolution
    \item Sensitivity $\rightarrow$ $\sim 0.469-0.942$N \cite{Fang2025Force}
    \item Dynamic range $\rightarrow$ $0-35$N \cite{Fang2025Force}
    \item Response time $\rightarrow$ $17$ms \cite{Xin2025Visionbased}
    \item Hysteresis
    \end{itemize}
  \\[-0.1\topsep]
  \bottomrule
  \end{tabular}
\end{table*}

\subsubsection{Commercial Sensors}
Artificial tactile sensors are much less established than artificial visual receptors such as RGB(-D) cameras and event-based cameras. Technologies for tactile sensing have been developed since the early 1970s and have undergone significant improvements in the past decade \citep{Dahiya2010Tactile, Dahiya2013Tactile, Kappassov2015Tactile}. However, the field remains young, with no widely accepted technical solutions and no unified tactile representation yet. Table~\ref{tab:transduction-mechanisms} summarizes the advantages and disadvantages of different transduction principles for detecting mechanical pressure. For additional information, please refer to \cite{Chi2018Recent,Meribout2024Tactile}.

Although several commercial solutions are available, costs remain relatively high, and performance is not always satisfactory for practical deployment. We therefore present representative commercial tactile sensing solutions. We are aware of other sensors such as the WTS-FT by Weiss Robotics GmbH \& Co.\ KG.\ and BioTac\textsuperscript{\textregistered} by SynTouch\textsuperscript{\textregistered}, but we include only devices that are still being produced and commercialized at the time of writing.

\begin{figure}[tb]
\centering
\begin{subfigure}[t]{\columnwidth}
\centering
\includegraphics[width=0.40\textwidth]{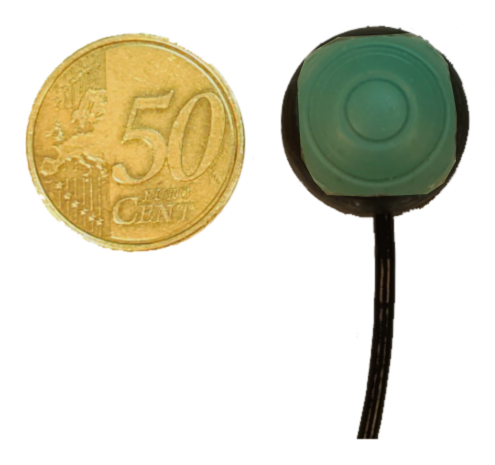}
\includegraphics[width=0.40\textwidth]{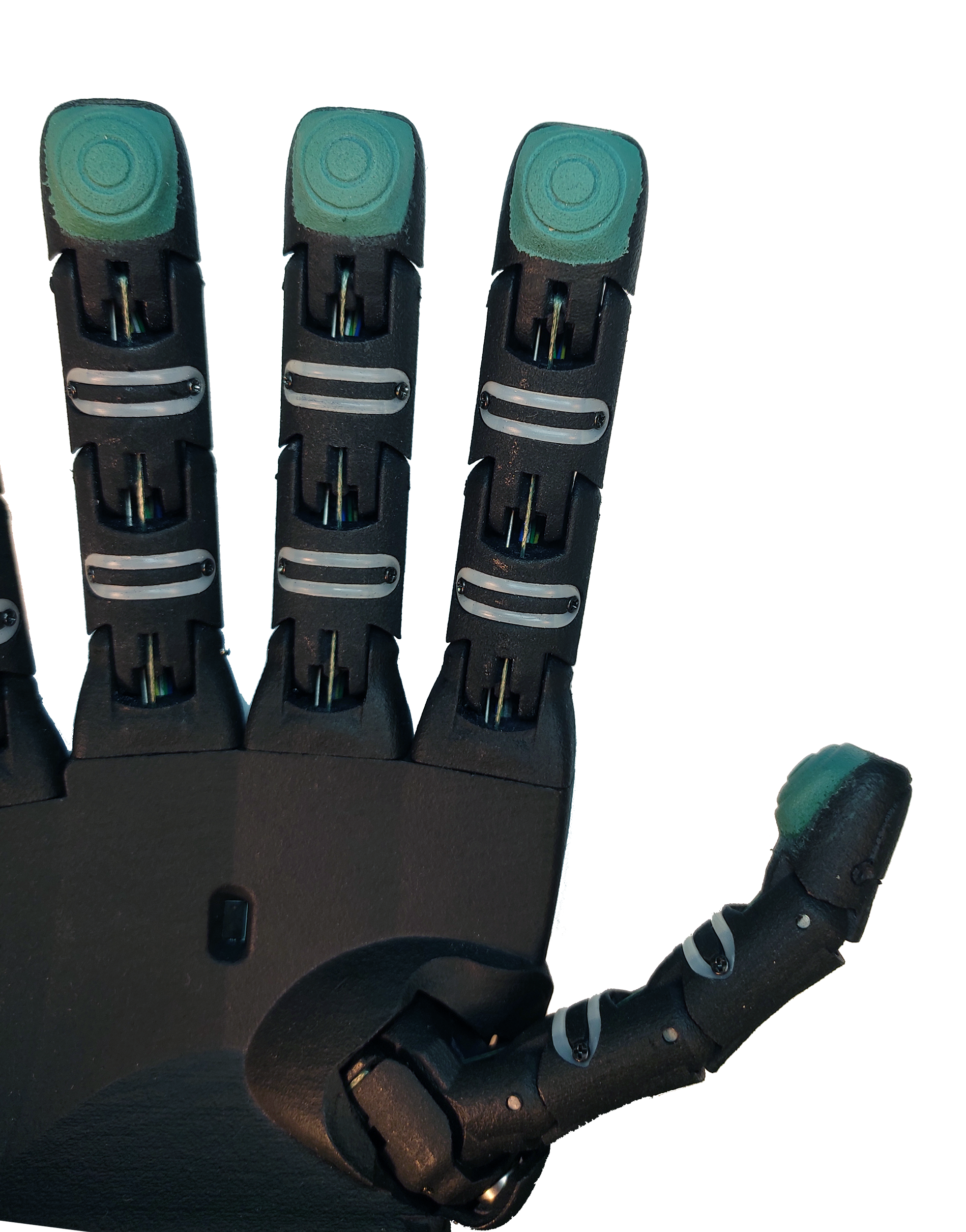}
\caption{Left: the SINGLEX stand-alone tactile pressure sensor version. Right: FTS tactile pressure sensor mounted on a robot finger. Images used with permission from Seed Robotics (\url{https://www.seedrobotics.com/}).}
\label{fig:seedrobotics}
\end{subfigure}
\hfill
\begin{subfigure}[t]{\columnwidth}
\centering
\includegraphics[width=0.40\textwidth]{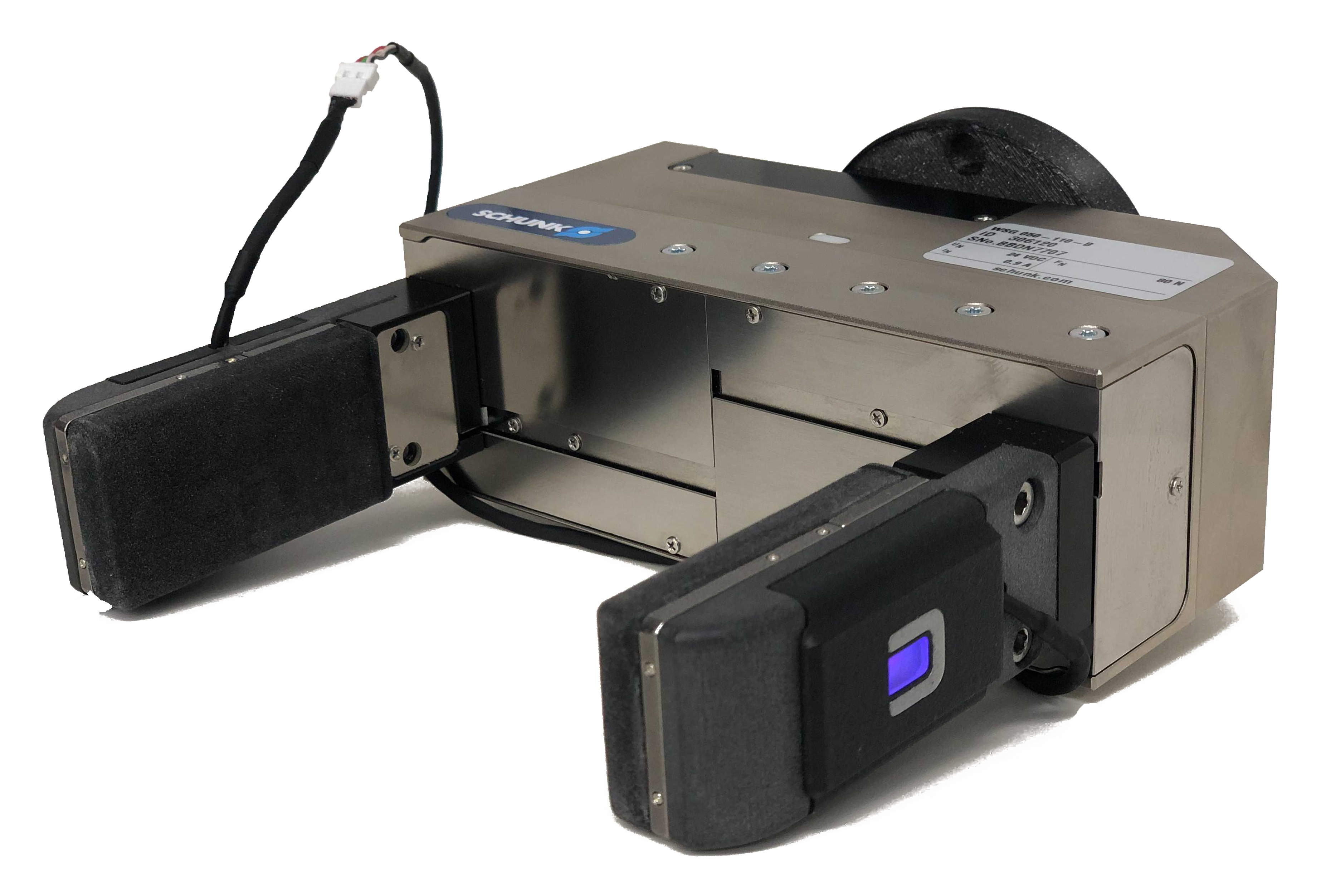}
\includegraphics[width=0.40\textwidth]{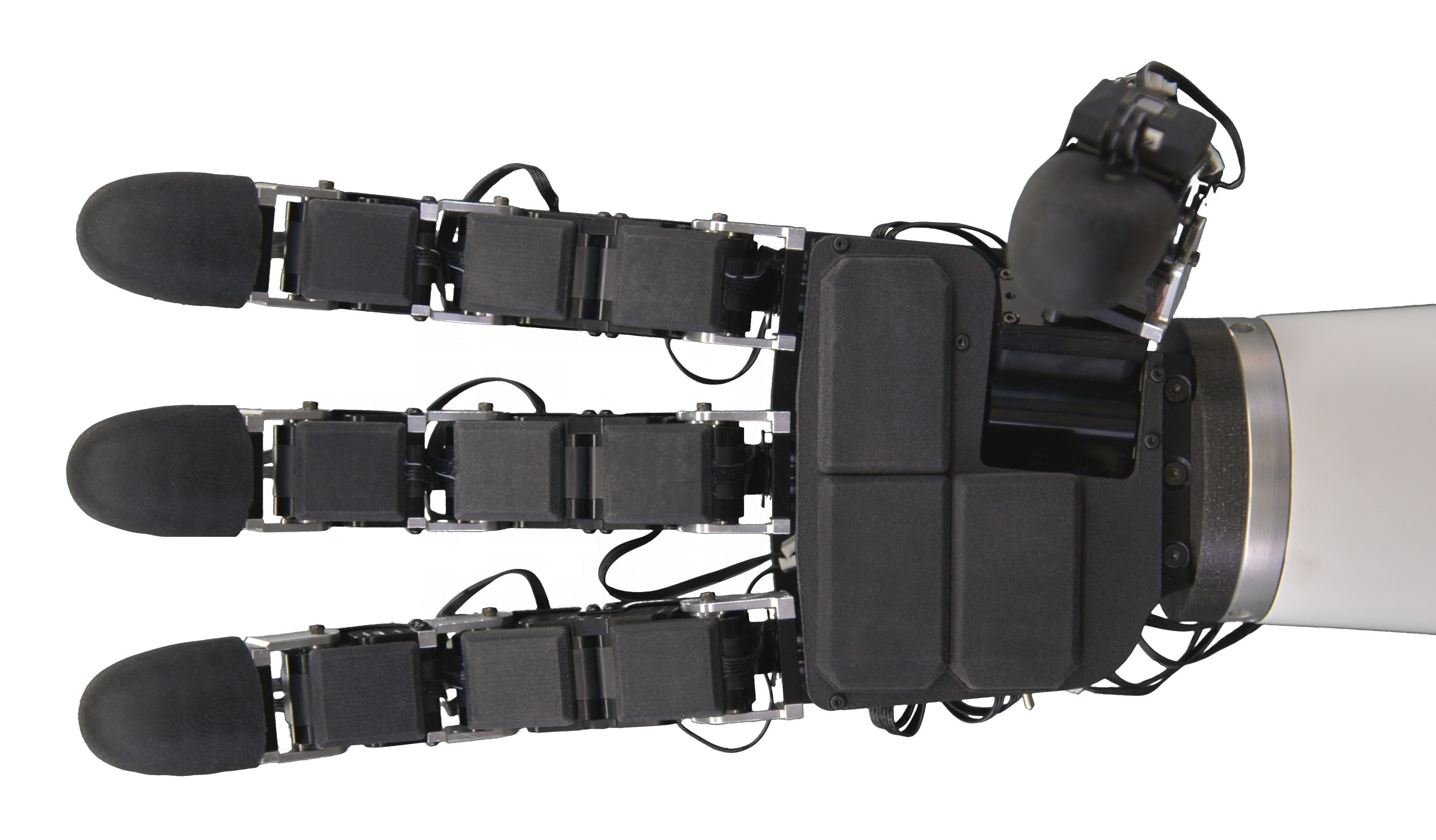}
\caption{Left: a flat version inspired by \cite{Tomo2018New}. Right: a multi-curved version inspired by \cite{Tomo2018Covering}. Images with permission from Xela Robotics (\url{https://xelarobotics.com/}).}
\label{fig:uskin}
\end{subfigure}
\hfill
\begin{subfigure}[t]{\columnwidth}
\centering
\includegraphics[width=0.48\textwidth]{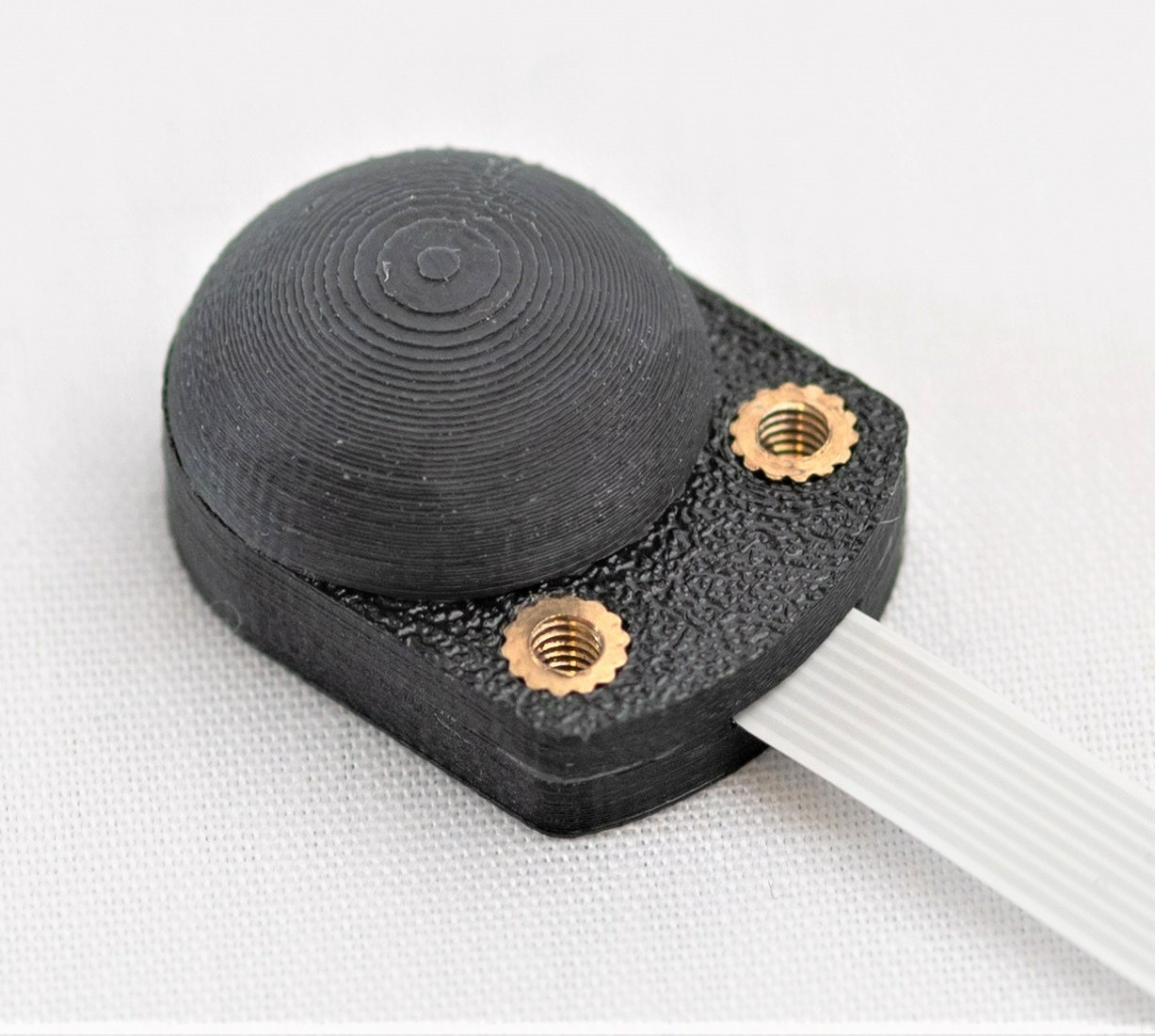}
\includegraphics[width=0.48\textwidth]{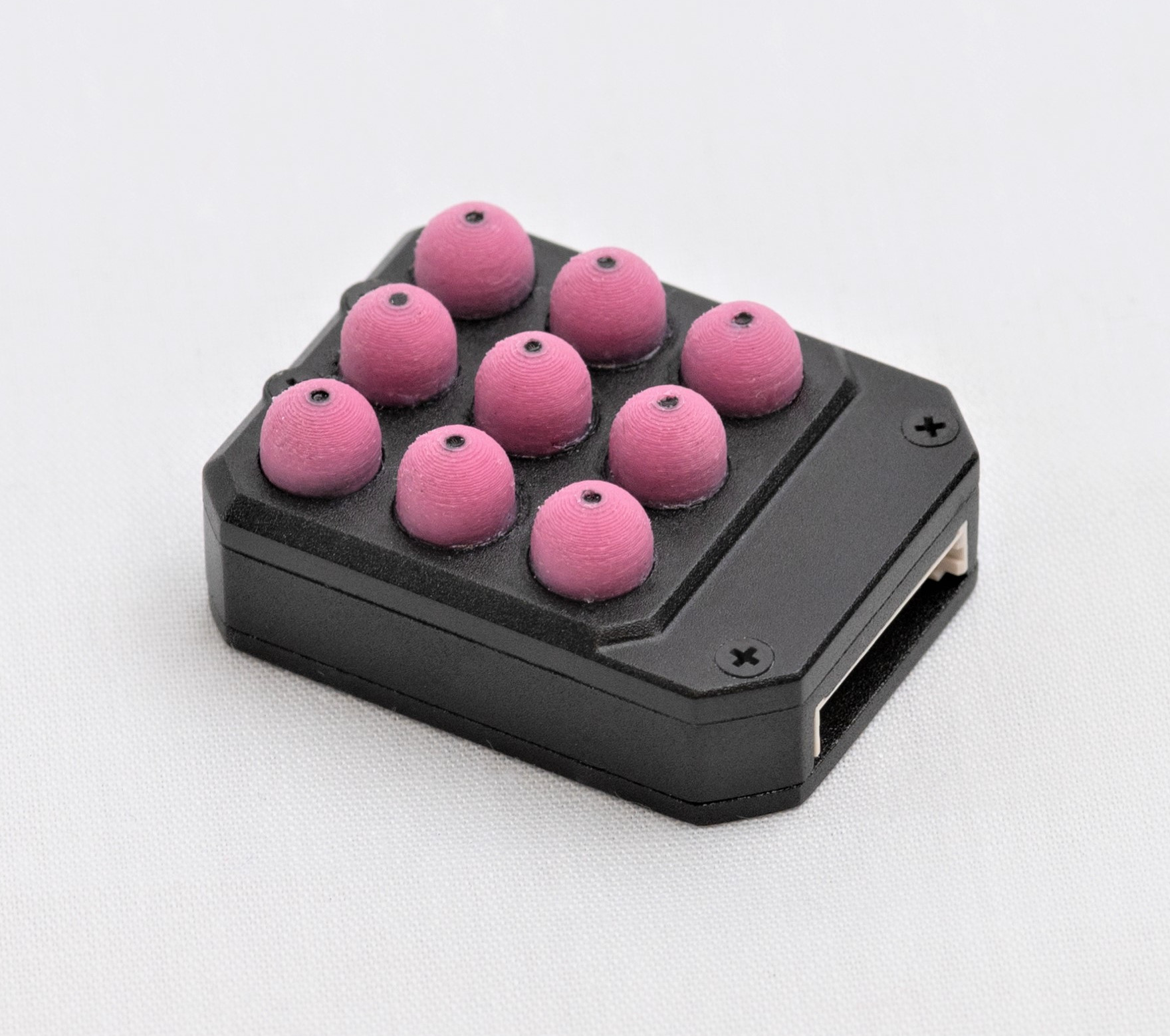}
\caption{Left: Single 3D force tactile sensor. Right: A slim tactile sensor array (PapillArray Sensor) available in different configurations. Images from Contactile (\url{https://contactile.com/}), licensed under CC BY-NC-ND 4.0.}
\label{fig:contactile}
\end{subfigure}
\caption{Tactile sensors.}
\label{fig:tactile_sensors}
\end{figure}

Seed Robotics' FTS tactile pressure sensors (see Fig.~\ref{fig:seedrobotics}) are low-cost sensors that provide high-resolution contact force measurements in the range of 1mN to 30N, with a sample frequency of 50Hz. The sensor compensates for temperature and is immune to magnetic interference, improving robustness in typical robotic environments.

The uSkin sensor \citep{Tomo2016Modular} by Xela Robotics is a magnetic tactile sensor composed of small magnets embedded in a thin layer of flexible rubber placed above a matrix of magnetic Hall-effect sensor chips. Upon contact, the magnets are displaced, and the magnetic field sensed by the Hall-effect chips changes, from which the contact forces are estimated. The uSkin sensor can measure the full 3D force vector, that is, both normal and shear contact forces, at each tactel, with good spatial resolution (about 1.6 tactels per square cm), high sensitivity (minimum detectable force of 1gf), and high sampling frequency ($>100$Hz, depending on the configuration). Different versions of the sensor are available to cover both flat and multi-curved surfaces, see Fig.~\ref{fig:uskin} for an example.

The GelSight Mini \citep{Yuan2017GelSight} and DIGIT \citep{Lambeta2020DIGIT} tactile sensors by GelSight are optical tactile sensors that utilize a piece of elastomeric gel with a reflective membrane coating on top, enabling them to capture fine geometrical textures as deformations in the gel. A series of RGB LEDs illuminates the gel, allowing a camera to record the deformation and output it at up to 30Hz.

Contactile offers both a stand-alone sensor and tactile sensor arrays, called PapillArray sensors, as shown in Fig.~\ref{fig:contactile}. These optical sensors consist of infrared LEDs, a diffuser, and four photodiodes encapsulated in a soft silicone membrane. The photodiodes measure light-intensity patterns, which are then used to infer the membrane's displacement and the force applied to it. This strategy enables the measurement of 3D deflections, 3D forces, and 3D vibrations, as well as the inference of emergent properties such as torque, incipient slip, and friction. The sensor can measure 3D applied forces up to 20N at a rate of up to 1kHz.

The availability of such commercial technologies is driving research forward in both the development of new sensing principles and applications such as robotic grasping, smart prostheses, and surgical robots. In particular, enhancements are still needed in mechanical robustness, sensitivity, and measurement reliability, as well as in ease of electromechanical integration and replacement, to enable wider deployment of tactile sensors in practical applications. Of particular interest are solutions that are flexible \citep{Larson2016Highly, SenthilKumar2019Review}, stretchable \citep{Buescher2015Flexible}, and able to cover sizeable \citep{Dahiya2013Directions} and curved \citep{Tomo2018Covering} surfaces, possibly with a small number of electrical connections \citep{ZaiElAmri2026VibroSense, JuinaQuilachamin2023Biomimetic}, can detect multiple contacts at the same time \citep{Hellebrekers2020Soft}, can detect both normal and shear forces \citep{Tomo2018New}, and are affordable and easy to manufacture \citep{Paulino2017LowCost}. For more information on experimental tactile sensing technologies, see \cite{Chi2018Recent}, and for a specialized review of printable, flexible, and stretchable tactile sensors, see \cite{SenthilKumar2019Review}.

From a perceptual perspective, human hands have a vast array of sensors, including mechanoreceptors, thermoreceptors, and nociceptors, which provide detailed information about touch, texture, temperature, and potential for damage (Section~\ref{sec:human_perception}). This sensory input is essential to perform a wide range of object manipulations, allowing humans to adjust their grip forces and hand or finger positions in real time. In contrast, robotic hands rely on a limited number of tactile or force sensors, which are often bulky relative to the robotic fingers and provide only partial information about the environment. Additionally, the data from these sensors requires post-processing to be interpreted and used effectively in manipulation tasks. These limitations suggest the need for further development in at least the mechanical and electrical design and signal-processing aspects of robotic perception sensors.

\section{Research Questions}
The motivation for this review is the idea that greater mechanical and sensory complexity, with anthropomorphic hands at the upper end of the complexity scale, lowers the barrier to broad skill coverage: a hand that mechanically enables more skill categories should require less task-specific engineering effort to reach each skill. The following research questions ask which specific features are responsible for this potential advantage, using the range of demonstrated skill repertoire as the outcome measure.

We expect manipulation capabilities to increase with the number of fingers.
Many human grasps can be performed with the thumb and an additional virtual finger~\cite{Montagnani2016Independent}, and almost all human grasp types can be realized with three fingers~\cite{Abbasi2016Grasp}.
Similarly, birds can grasp and adjust their grasps, but parrots use their tongues to realize one in-hand manipulation DoF with three virtual fingers~\cite{sugasawa2021object}. Individuals with six fully functional fingers exhibit better manipulation capabilities than individuals with five fingers~\cite{Mehring2019Augmented}.
\vspace{-6pt}
\begin{mdframed}[backgroundcolor=gray!20, linewidth=0pt]
RQ1: Does the number of fingers extend a robotic hand's skill repertoire? -- Hypothesis: Skill repertoire size and / or manipulated in-hand DoF increase with the number of fingers.
\end{mdframed}

In comparison to monkeys, human fingers move more independently~\cite{Hager-Ross2000Quantifying}. This suggests that higher individuation of finger movements is beneficial for manipulation.
\vspace{-9pt}
\begin{mdframed}[backgroundcolor=gray!20, linewidth=0pt]
RQ2: Do more DoF and actuators extend a robotic hand's skill repertoire?\\
Hypothesis: Skill repertoire size and / or manipulated in-hand DoF increase with the number of DoF and actuators.
\end{mdframed}

Precision grasps of the human hand with small contact areas rely more on independent abduction/adduction of the fingers than on independent flexion/extension, because this allows for a precise opposition of the thumb~\cite{Montagnani2016Independent}.
\vspace{-9pt}
\begin{mdframed}[backgroundcolor=gray!20, linewidth=0pt]
RQ3: Do abduction/adduction joints extend a robotic hand's skill repertoire? -- Hypothesis: Hands with abduction/adduction joints demonstrate a larger skill repertoire than those without.
\end{mdframed}

In humans, cutaneous receptors are essential for adjusting grip forces to object weight and surface properties, detecting tangential forces and incipient slip, segmenting movements into distinct grasp phases, and timing sequential actions such as playing an instrument, while the absence or degradation of tactile feedback leads to prolonged grasp phases, impaired coordination, and unstable grasps even when vision is available \citep{Dahiya2010Tactile,Johansson2009Coding}. The combination of tactile and proprioceptive feedback further allows humans to adapt to varying textures, weights, and shapes in dynamic environments and to recognize and differentiate objects when visual information is limited \citep{Lederman2009Haptic}. Taken together, these findings suggest that systems that actively use tactile feedback for control can support more varied and adaptable manipulation behaviors than systems that rely on kinematics and vision alone \citep{Lederman2009Haptic,Dahiya2010Tactile,sugasawa2021object}.
\vspace{-9pt}
\begin{mdframed}[backgroundcolor=gray!20, linewidth=0pt]
RQ4: Does tactile sensing extend a robotic hand's skill repertoire? -- Hypothesis: Hands with tactile sensing demonstrate a larger skill repertoire than those without.
\end{mdframed}\vskip -0.8cm
\begin{mdframed}[backgroundcolor=gray!20, linewidth=0pt]
RQ5: Does richer sensing extend a robotic hand's skill repertoire? -- Hypothesis: Skill repertoire size increases with the number of sensor modalities.
\end{mdframed}

In addition, we want to answer more general questions about the state of research on robotic hand design and robotic manipulation skills to put the results into context.
\vspace{-9pt}
\begin{mdframed}[backgroundcolor=gray!20, linewidth=0pt]
RQ6: How complex are the hands used in current manipulation research compared to the human hand?
\end{mdframed}\vskip -0.8cm
\begin{mdframed}[backgroundcolor=gray!20, linewidth=0pt]
RQ7: How thoroughly has the skill repertoire of robotic hands been explored?
\end{mdframed}

The systematic review assesses the skill repertoire size according to the taxonomy introduced by Dollar \cite{Dollar2014Classifying}. Moreover, this taxonomy could guide the creation of a standard task set for each leaf, providing a structured basis for comparing robot hand dexterity. The following analysis is a direct implementation of that proposal. The taxonomy contains two in-hand manipulation categories for which we investigate the manipulated DoF of the objects.

\section{Systematic Review of Robotic Hands and their Skill Repertoire}
We performed a systematic literature review to connect hands with the skills they can perform. We follow the PRISMA Statement~\cite{Moher2009Preferred} for reporting but deviate from it, since we do not aggregate the results of empirical studies with statistical results. We only collect information on demonstrated robot skills as there is a lack of standardization of design goals or requirements, and, more importantly, testing protocols.

\subsection{Methods}
\subsubsection{Search Strategy}
We extracted information about which hand was used to perform which skills.
For this purpose, on May $7^{th}$ 2025, we searched Google Scholar and IEEE Xplore with the search term \textit{(hand or gripper) and (manipulation or grasping or grip or skill)}.
We selected both databases because they are complementary and have an API to programmatically filter records. Google Scholar is one of the largest overall scientific databases indexing scientific articles, preprints, and university websites. IEEE Xplore is one of the main publishers in engineering and robotics research.
We extracted the title, number of citations, year of publication, source (e.g., journal, conference), publisher, text snippet, and URL from Google Scholar. We extracted the title, number of citations, year of publication, source, abstract, URL, and keywords from IEEE Xplore. These records were filtered and selected in the following step.

\subsubsection{Selection}
Fig.~\ref{fig:study_selection} summarizes the selection process.

\paragraph{Automatic filtering}
We filtered the results from Google Scholar and IEEE Xplore using the following criteria: publication dates must be between 2019 and 2025, the article must have at least one citation, be at least four pages long, and be written in English. All IEEE papers were removed from the initial set retrieved using Google Scholar to eliminate duplicates. All non-peer-reviewed articles published on arXiv were removed from our list.
Finally, we filtered IEEE papers based on the keyword terms to ensure they contained one of the following terms: (hand or gripper) and (manipulation or grasping or grip or skill).

\paragraph{Manual screening of records}
In a manual screening step, we excluded papers not related to robotic or prosthetic hands or grippers based on the title, abstract, or text snippet. These often involve measurements of hand grip strength in humans or other topics in the medical field.
Each record was screened by at least one author.

\paragraph{Assessment of full-text articles}
After automatic filtering and manual prescreening, we kept papers demonstrating skills with multi-fingered hands ($\geq 2$ fingers) beyond simple pinch grasps. We excluded papers with the following criteria.
\begin{itemize}[leftmargin=*,noitemsep,nolistsep]
\item Virtual reality / augmented reality
\item Gripper / hand design / development without demonstration of skills beyond flexion/extension of fingers
\item Development of sensors or perception approaches without demonstration of skills beyond flexion/extension of fingers
\item Non-anthropomorphic gripper designs, such as soft fingers or suction mechanisms
\end{itemize}
Each record was screened by one author. Articles that were difficult to classify were discussed among all authors.

\subsubsection{Data Collection Process}
For each article, we extracted the hand and its features, sensor setup, and skill categories that were presented or are trivially possible to implement with the hand shown in the article.
Data were extracted by one author; ambiguous cases were discussed among all authors.

\paragraph{Hand features}
We extracted the following features, all defining characteristics of the human hand, from the articles:
number of fingers, DoF, number of actuators, mechanism features (use of the palm, thumb reposition/opposition joint, abduction/adduction joints, flexion/extension joints), sensor features (tactile / kinesthetic / visual feedback used or not).

\paragraph{Skill repertoire}
We examined the papers and extracted any skills or movements that the authors claim to have demonstrated for a real or simulated robotic hand.
These manipulation skills were categorized according to the taxonomy of manipulation tasks by Dollar~\cite{Dollar2014Classifying}, using the example skill Dollar provides for each leaf of the taxonomy (Fig.~5 in \cite{Dollar2014Classifying}) as the representative task for that category or an equivalent demonstrated skill.
The taxonomy encompasses both prehensile skills, such as grasping and in-hand manipulation, and non-prehensile skills, including pushing and gestures.
It is designed to be independent of the hand morphology.
Furthermore, for in-hand manipulation categories (14 and 15: Contact / Motion / Within Hand / (No) Motion at Contact), we extract the number of manipulated DoF of the object.

A skill is considered possible without demonstration in a few cases: (1) all hands are credited with category~1 (rest position); (2) all hands that can form a surface with a flat palm and its fingers are credited category~4 (open handed hold), since this provides a sufficiently large support surface, (3) all hands are credited with category~10 (holding object still), since every hand in the dataset provides at least two opposing contact points, which is sufficient for a grasp without object motion; (4) all hands with tactile feedback that can adapt to the shape of a cylindrical object were credited category~11 (sliding hand along handrail), since it is sufficient to have a tactile sensor to not apply too much force from both directions to slide along the handrail.

\subsubsection{Risk of Bias in Individual Studies}
We see the risk of underreported skill repertoires for the robotic hands.
Not all hands fully exhaust their skill set in their respective publication. Even though we aggregate results from various publications using the same hand, we cannot be sure that the hand would not be able to perform an in-hand manipulation skill with more DoF or another category of skills not seen in the publications.
Thus, it is likely that some skills that a hand can perform are not investigated in the respective publication. Users of prosthetic hands may be able to perform a large variety of skills in their daily lives that are not reported in any publication.
Hence, we analyze extreme cases, e.g., the minimum complexity of a hand required to implement a specific skill.

Another limitation is the binary nature of reporting skills, which means it is not possible to infer the quality or proficiency of performing a skill with a particular hand.

\subsubsection{Summary Measures and Synthesis of Results}
If a hand is used in multiple papers, we aggregate results by taking the union of demonstrated skills (logical OR across papers). Analyses are conducted both per paper-hand combination ($N=128$) and per hand after aggregation ($N=93$).

All statistical methods were selected, all results were verified, and all interpretations were done by the authors. The statistical analysis implementation was done with the assistance of Claude Code (model: Sonnet-4.6, Anthropic).

\paragraph{Correlation analysis}
All statistical analyses were implemented in Python~3.12 using NumPy~2.4.4, SciPy~1.17.1, and statsmodels~0.14.6.
We measure correlation with the following methods.
\textit{(i) Spearman $\rho$} with 95\,\% bootstrap CI (5000 resamples): count features (fingers, DoF, actuators, sensor count) against in-hand DoF (cat. 14 / cat. 15, RQ1/RQ2) and skill repertoire size (RQ1/RQ2/RQ5). Statistical significance tested with $t = \rho \cdot \sqrt{N-2} / \sqrt{1 - \rho^2}$, $df = N-2$.
\textit{(ii) Mann-Whitney U / rank-biserial correlation $r_{rb}$} \cite{Wendt1972Dealing} with 95\,\% bootstrap CI (5000 resamples): abduction/adduction against in-hand DoF (cat. 14 / cat. 15, RQ3) and binary features (abduction/adduction, tactile feedback, kinesthetic feedback) against skill repertoire size (RQ3/RQ4). Statistical significance tested with one-sided Mann-Whitney U.
All 15 tests are corrected for multiple comparisons using the Benjamini--Hochberg false discovery rate procedure ($\alpha=0.05$) \cite{Benjamini1995Controlling}.

\paragraph{Analysis of extreme cases}
We plot the distribution of hand features that define the hand's complexity versus skill complexity to identify extreme cases.
More specifically, we look for simple hands that can perform complex skills. We decided to introduce this measure post-hoc after seeing the results of the correlation analysis.

\paragraph{Histogram of hand and sensor features}
To get an overview of the distribution of hand and sensor features of the presented hands, we present histograms of these features.

\paragraph{Percentage of achieved skills by hands}
We use bar charts to show the percentage of hands with specific features that demonstrate the ability to perform each skill category. Although in theory, each skill category should be covered by each hand, this analysis gives us an impression of how thoroughly the skill repertoire has been examined.

\subsection{Results}

\subsubsection{Selection}
Fig.~\ref{fig:study_selection} summarizes the selection process.
We include 125 papers in the summary.
See supplementary material for an overview of included articles, \url{https://github.com/AlexanderFabisch/JIRS-HandsSurvey}.

\begin{figure}[bt]
\centering
\includegraphics[width=\columnwidth]{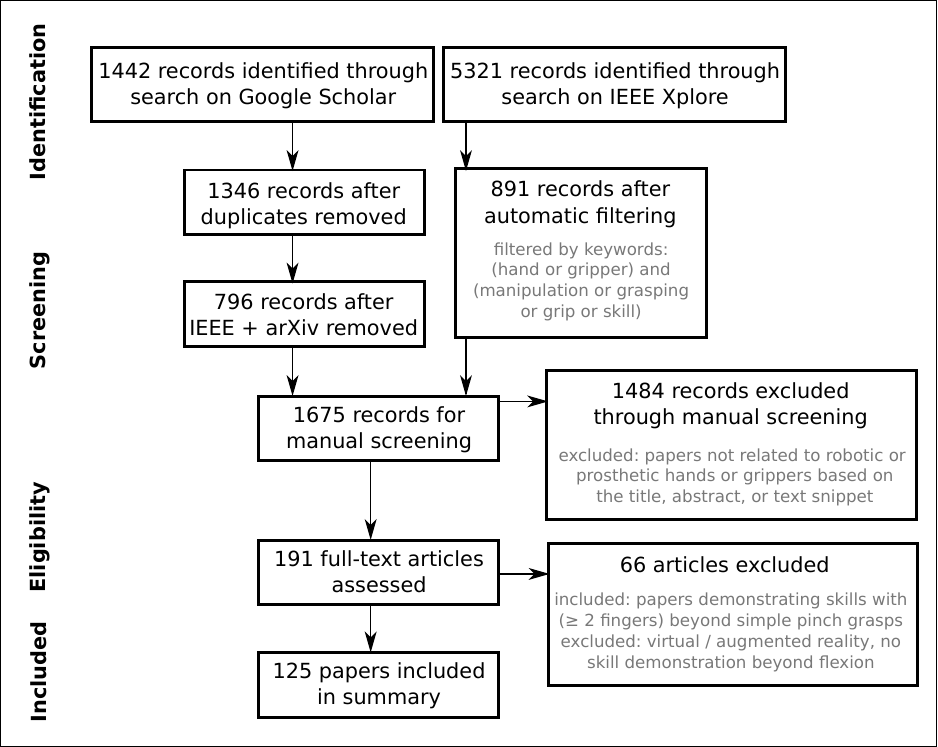}
\caption{Summary of selection process.}
\label{fig:study_selection}
\end{figure}

\subsubsection{Synthesis of Results}

\paragraph{Correlation analysis}
Tables~\ref{tab:correlation_analysis} and~\ref{tab:mannwhitney} summarize the results. All $p^*$ values are FDR-corrected (Benjamini--Hochberg, $\alpha=0.05$); significant results are shown in bold.

\begin{table*}[!t]
\footnotesize
\setlength{\tabcolsep}{3pt}
\renewcommand{\arraystretch}{1.2}
\begin{minipage}[t]{\columnwidth}
\centering
\captionof{table}{Spearman $\rho$ for mechanism and sensor features vs.\ in-hand manipulation DoF and skill repertoire size. Level: PP\,=\,per-paper-hand ($N_\text{PP}$), PH\,=\,per-hand ($N_\text{PH}$). $p^*$: FDR-corrected $p$-value (Benjamini--Hochberg, $\alpha=0.05$); bold and starred entries are significant.}
\label{tab:correlation_analysis}
\begin{tabular}{@{}lllcccc@{}}
\toprule
RQ & Feature & Lvl & $N$ & $\rho$ & 95\,\% CI & $p^*$ \\
\midrule
\multicolumn{7}{@{}l}{\textit{Mechanism features vs.\ in-hand DoF}} \\[2pt]
\multicolumn{7}{@{}l}{\quad\textit{Motion DoF}} \\[1pt]
RQ1 & \# Fingers & PP & 40 & $+0.074$ & $[-0.24,\,+0.37]$ & 0.885 \\
 & & PH & 31 & $+0.277$ & $[-0.09,\,+0.61]$ & 0.178 \\[2pt]
RQ2 & DoF & PP & 40 & $-0.002$ & $[-0.36,\,+0.34]$ & 0.992 \\
 & & PH & 31 & $+0.357$ & $[-0.04,\,+0.67]$ & 0.081 \\[2pt]
RQ2 & \# Actuators & PP & 39 & $+0.043$ & $[-0.31,\,+0.37]$ & 0.964 \\
 & & PH & 30 & $+0.396$ & $[+0.04,\,+0.67]$ & 0.057 \\[2pt]
\multicolumn{7}{@{}l}{\quad\textit{No-Motion DoF}} \\[1pt]
RQ1 & \# Fingers & PP & 17 & $+0.029$ & $[-0.50,\,+0.48]$ & 0.978 \\
 & & PH & 15 & $+0.066$ & $[-0.50,\,+0.53]$ & 0.875 \\[2pt]
RQ2 & DoF & PP & 17 & $+0.239$ & $[-0.29,\,+0.69]$ & 0.593 \\
 & & PH & 15 & $+0.288$ & $[-0.31,\,+0.76]$ & 0.343 \\[2pt]
RQ2 & \# Actuators & PP & 16 & $+0.056$ & $[-0.48,\,+0.58]$ & 0.964 \\
 & & PH & 14 & $+0.033$ & $[-0.58,\,+0.61]$ & 0.912 \\[2pt]
\midrule
\multicolumn{7}{@{}l}{\textit{Features vs.\ skill repertoire size}} \\[2pt]
RQ1 & \# Fingers & PP & 128 & $+0.449$ & $[+0.29,\,+0.59]$ & \textbf{$<$0.001}$^*$ \\
 & & PH & 93 & $+0.437$ & $[+0.25,\,+0.61]$ & \textbf{$<$0.001}$^*$ \\[2pt]
RQ2 & DoF & PP & 127 & $+0.541$ & $[+0.40,\,+0.66]$ & \textbf{$<$0.001}$^*$ \\
 & & PH & 93 & $+0.500$ & $[+0.34,\,+0.64]$ & \textbf{$<$0.001}$^*$ \\[2pt]
RQ2 & \# Actuators & PP & 123 & $+0.498$ & $[+0.35,\,+0.63]$ & \textbf{$<$0.001}$^*$ \\
 & & PH & 89 & $+0.365$ & $[+0.15,\,+0.55]$ & \textbf{0.001}$^*$ \\[2pt]
RQ5 & \# Sensors & PP & 128 & $+0.350$ & $[+0.18,\,+0.50]$ & \textbf{$<$0.001}$^*$ \\
 & & PH & 93 & $+0.274$ & $[+0.07,\,+0.46]$ & \textbf{0.017}$^*$ \\[2pt]
\bottomrule
\end{tabular}
\end{minipage}\hfill
\begin{minipage}[t]{\columnwidth}
\centering
\captionof{table}{Mann--Whitney $U$ test: binary hand features vs.\ outcome variables. $r_\text{rb}$: rank-biserial correlation (effect size). Level: PP\,=\,per-paper-hand ($N^+_\text{PP}$), PH\,=\,per-hand ($N^+_\text{PH}$). $p^*$: FDR-corrected $p$-value (Benjamini--Hochberg, $\alpha=0.05$); bold and starred entries are significant.}
\label{tab:mannwhitney}
\begin{tabular}{@{}lllccc@{}}
\toprule
RQ & Feature & Lvl & $N^+$ & $r_\text{rb}$ [95\,\% CI] & $p^*$ \\
\midrule
\multicolumn{6}{@{}l}{\textit{Binary features vs.\ Skill rep.\ size}} \\[2pt]
RQ3 & Abd./Add. & PP & 57 & $+0.464$ $[+0.29,\,+0.63]$ & \textbf{$<$0.001}$^*$ \\
 & & PH & 30 & $+0.481$ $[+0.26,\,+0.68]$ & \textbf{$<$0.001}$^*$ \\[2pt]
RQ4 & Tactile & PP & 18 & $+0.491$ $[+0.23,\,+0.72]$ & \textbf{$<$0.001}$^*$ \\
 & & PH & 13 & $+0.445$ $[+0.13,\,+0.73]$ & \textbf{0.012}$^*$ \\[2pt]
RQ4 & Kinesthetic & PP & 62 & $+0.501$ $[+0.33,\,+0.66]$ & \textbf{$<$0.001}$^*$ \\
 & & PH & 41 & $+0.438$ $[+0.23,\,+0.64]$ & \textbf{$<$0.001}$^*$ \\[2pt]
\midrule
\multicolumn{6}{@{}l}{\textit{Binary features vs.\ Motion DoF}} \\[2pt]
RQ3 & Abd./Add. & PP & 17 & $+0.003$ $[-0.36,\,+0.37]$ & 0.750 \\
 & & PH & 9 & $+0.318$ $[-0.12,\,+0.71]$ & 0.120 \\[2pt]
\midrule
\multicolumn{6}{@{}l}{\textit{Binary features vs.\ No-Motion DoF}} \\[2pt]
RQ3 & Abd./Add. & PP & 11 & $+0.242$ $[-0.24,\,+0.73]$ & 0.381 \\
 & & PH & 9 & $+0.278$ $[-0.24,\,+0.74]$ & 0.225 \\[2pt]
\bottomrule
\end{tabular}
\end{minipage}
\end{table*}

\begin{figure*}[tb]
\centering
\includegraphics[width=\textwidth]{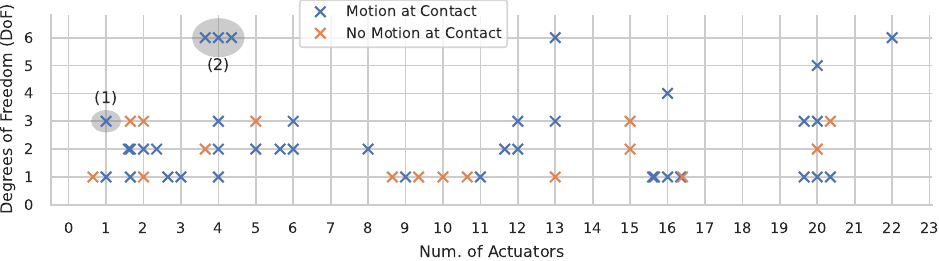}
\caption{Extreme case analysis in number of actuators versus degrees of freedom that could be controlled in one of the in-hand manipulation categories. Extreme cases: (1) 3 controllable DoF of the object: Chavan-Dafle et al.\ \cite{Chavan-Dafle2020Planar} with one actuator (2 fingers, 2 DoF); (2) 6 controllable DoF of the object: two hands of Liarokapis and Dollar \cite{Liarokapis2019Combining} with four actuators (3 fingers, 7 DoF; 4 fingers, 8 DoF), and Morgan et al.\ \cite{Morgan2022Complex} with four actuators (4 fingers, 8 DoF).}
\label{fig:corr_doa}
\vskip -0.8cm
\end{figure*}

\paragraph{Analysis of extreme cases}
Outliers that we identified in Fig.~\ref{fig:corr_doa} are (1) Chavan-Dafle et al.\ \cite{Chavan-Dafle2020Planar}
with 2 fingers, 2 DoF, 1 actuator, and 3 controllable DoF of the object, as well as (2) the following examples of 6 controllable DoF of the object: two hands of Liarokapis and Dollar \cite{Liarokapis2019Combining} with 3 fingers, 7 DoF, 4 actuators, and 4 fingers, 8 DoF, 4 actuators respectively, and of Morgan et al.\ \cite{Morgan2022Complex} with 4 fingers, 8 DoF, 4 actuators.

\begin{figure*}[tb]
\centering
\includegraphics[width=\textwidth]{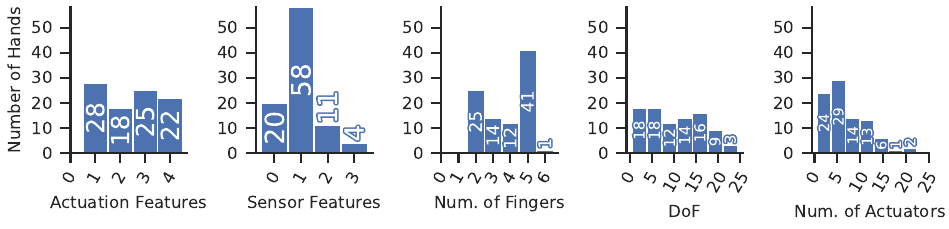}
\caption{Histograms (max.\ of 8 bins) of the characteristics of the hands that we analyzed.}
\label{fig:hand_features}
\vskip -0.4cm
\end{figure*}

\paragraph{Histogram of hand and sensor features}
Fig.~\ref{fig:hand_features} shows a summary of the hand characteristics for all hands in the analyzed papers.

\begin{figure*}[tb]
\centering
\includegraphics[width=\textwidth]{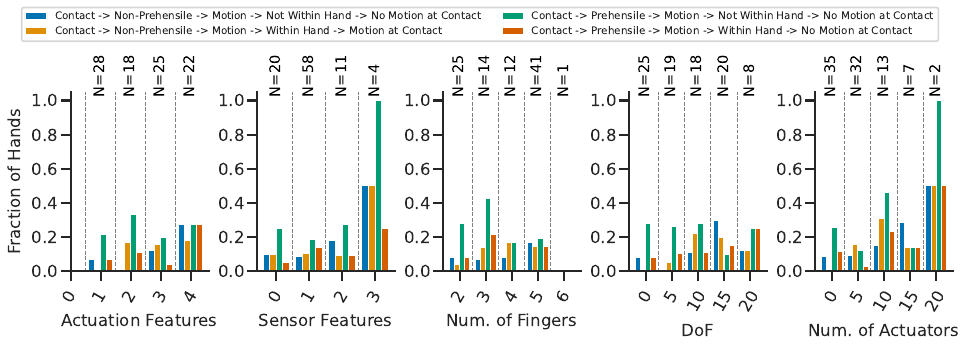}
\caption{Percentage of hands that were shown to perform a skill from the respective skill category.}
\label{fig:skill_repertoire}
\vskip -0.4cm
\end{figure*}

\paragraph{Percentage of achieved skills by hands}
Fig.~\ref{fig:skill_repertoire} shows the percentage of hands that can perform a skill from the respective skill category, ordered by features of the hands. Due to space constraints, we only show the result for four skill categories of Dollar's taxonomy~\cite{Dollar2014Classifying}: category~6 (e.g., pushing a coin), category~9 (e.g., rolling a ball on a table), category~12 (e.g., turning a doorknob), and category~14 (e.g., writing). These skill categories represent a wide range of different skills.

\subsection{Discussion}

\subsubsection{Summary of Evidence}

\paragraph{Complexity does not predict in-hand manipulation depth}
No hand complexity metric correlates significantly with the number of object DoF controlled during in-hand manipulation (Table~\ref{tab:correlation_analysis}), for either Motion-at-Contact or No-Motion-at-Contact skills. The null result holds even for the most complex hands, which were rarely pushed to their mechanical limits in the analyzed studies.

\paragraph{Complexity predicts skill breadth, not depth}
By contrast, mechanism complexity is a medium-strength predictor of skill repertoire breadth (Tables~\ref{tab:correlation_analysis} and~\ref{tab:mannwhitney}): all three count features (fingers, DoF, and actuators) show significant positive Spearman correlations, and hands with abduction/adduction joints, kinesthetic sensing, or tactile feedback demonstrate significantly broader repertoires. More capable hands are thus applied across a wider range of task categories, even if no single study exhausts their potential.

\paragraph{Simple mechanisms can achieve high in-hand manipulation DoF}

The extreme-case analysis (Fig.~\ref{fig:corr_doa}) reveals that simple, non-anthropomorphic mechanisms can match or exceed the in-hand DoF of complex hands, either by exploiting environmental contacts as a virtual finger \cite{Chavan-Dafle2020Planar} or by arranging four fingers in a symmetric opposition \cite{Morgan2022Complex,Liarokapis2019Combining}. Intelligent mechanism design can therefore compensate for low mechanical complexity.

\paragraph{Underexplored skill repertoire}
Fig.~\ref{fig:hand_features} shows the field is split between five-fingered anthropomorphic designs and minimal two-finger grippers, with other morphologies underexplored and sensor integration sparse (modal sensor count: one).
Fig.~\ref{fig:skill_repertoire} shows that regardless of hand type, the evaluated skill repertoire is narrow: in-hand manipulation categories are rarely tested, and when they are, only a few object DoF are controlled.
Measured repertoire sizes therefore systematically underestimate true capability, which biases the reported complexity--repertoire correlations toward zero.

\subsubsection{Limitations}
The range and complexity of possible manipulation tasks are almost impossible to quantify. While we have seen that simple hand mechanisms can control many DoF of an object, we cannot say that there is no better way of measuring task complexity, in which these mechanisms are worse than anthropomorphic hands. A better approach could be to measure the range of motion and other skill-specific performance criteria. We are unable to extract these from all publications in a comparable manner.
There is also no notion of complexity in the other skill categories that we analyzed (e.g., signing could be \textit{thumbs up} or complex sign language).
The possibility that hands appearing in more papers accumulate more skill annotations (a publication-scope confound) was assessed by aggregating results per hand; however, the correlations between hand complexity and skill repertoire size persist after aggregation ($N=93$).

\subsubsection{Conclusions}
Current robotic hands fall short of human-level sensor integration and control, and the skill repertoire of most hands is underexplored, underscoring the need for standardized evaluation protocols that cover the full taxonomy.

The null result for in-hand manipulation DoF arises from two complementary causes: complex hands rarely exhaust their manipulation capabilities in practice (most studies demonstrate only 1--3 object DoF), and simple non-anthropomorphic mechanisms can achieve equally high in-hand DoF by exploiting environmental contacts \citep{Chavan-Dafle2020Planar} or symmetric four-finger opposition \citep{Morgan2022Complex,Liarokapis2019Combining}.

Nevertheless, mechanism complexity is a significant predictor of skill repertoire breadth (Tables~\ref{tab:correlation_analysis} and~\ref{tab:mannwhitney}), suggesting that more capable hands are deployed across a wider range of task categories, even though no single study exhausts their potential.

\section{Discussion}
We compared the mechanical and sensory features of the human hand with those of robotic hands to address the question of whether robotic hands should mimic the human hand. In this section, we summarize and interpret the results and attempt to answer this question.

\subsection{Robotic Hand Mechanisms Achieve Complexity Similar to the Human Hand}
While robots are still inferior to humans in terms of sensory feedback, control, and learning, the case is not as clear-cut in terms of mechanical features.
Two five-fingered hands reach an amount of DoF and actuators that is comparable to the human hand: the Shadow Dexterous Hand and the bio-inspired hand of Zhou et al.~\cite{Zhou2021bio}. Both support abduction/adduction of fingers 2--5 and reposition/opposition of the thumb. The Shadow Dexterous Hand allows additional light opposition of the little finger, similar to the human hand. The hand of Zhou et al.~\cite{Zhou2021bio} allows palm folding, similar to the human hand. There are also 22 manipulation systems that integrate all four sensor modalities that we evaluated. Specifically, the Shadow Dexterous Hand can be combined with tactile sensors, replicating at least a considerable fraction of the complexity of the human hand in terms of sensor integration and mechanical complexity.

\subsection{The Skill Repertoire of Most Hands is Underexplored}
Even complex hands are evaluated in only a few skill categories of Dollar's taxonomy. Hence, measured skill repertoire sizes systematically underestimate each hand's actual capability.
To make hands, mechanisms, or complete manipulation systems, including sensors and control software, comparable, we need better evaluation protocols or benchmarks.

As pointed out earlier, Dollar~\cite{Dollar2014Classifying} proposed that his taxonomy could guide the creation of a standard task set for each skill category, and that each category should include tasks of varying difficulty. Building on this idea, evaluations should distinguish between prosthetic scenarios, in which human intelligence and sensory systems control the hand, and robotic scenarios, in which a complete manipulation system operates autonomously based only on its own sensors. Moreover, evaluation protocols should quantify performance proficiency rather than only task completion, for example, by including time to completion, degree of completion, and potentially additional task-specific metrics.

\subsection{In-Hand Manipulation Does Not Benefit from Anthropomorphic Hand Mechanisms}
Similarly, most mechanically complex hands in our systematic review do not exhaust their in-hand manipulation capabilities. At the same time, we found that in-hand manipulation tasks can be solved by simple mechanisms.

Chavan-Dafle et al.\ \cite{Chavan-Dafle2020Planar} demonstrate that a gripper with two fingers and one actuator can perform in-hand manipulation with 3 DoF by using the environment as a third contact surface in a clever way. Liang et al.\ \cite{Liang2024Robust} build upon this idea.
Similarly, humans constantly exploit environment constraints \cite{DellaSantina2017Postural}.
For this to be possible, robotic hands should be robust, i.e., flexible and soft, with complexity shifting to intelligent motor control and sensor integration.

Even without using the environment, there are several two-finger grippers that perform simple in-hand manipulation such as rolling an object (e.g.,~\cite{Bircher2021Complex,Xiang2024Adaptive}).

Only four actuators with three fingers or two pairs of opposing fingers are sufficient to manipulate all six DoF of an object~\cite{Liarokapis2019Combining,Morgan2022Complex}. The proposed circular finger arrangement is not anthropomorphic, but may be even better suited for many in-hand manipulation tasks. This is compatible with the result of Feix et al.~\cite{Feix2021Effect}, who show in human experiments that more than three fingers in contact with the object can interfere with each other.

\subsection{Mechanism Complexity Correlates With Skill Repertoire Size}
The central tension of our results is that mechanical complexity correlates with skill breadth ($\rho \approx .37$--$.50$, confirmed for fingers, DoF, actuators, and abduction/adduction) but not with in-hand manipulation depth (DoF controlled: all correlations non-significant).

Montagnani et al.\ \cite{Montagnani2016Independent} showed that independent abduction/adduction enables precision grasps with small contact areas in the human hand, and we found that it increases skill repertoire size significantly ($r_{rb} = .48$, $p^* < .001$). Similarly, more fingers ($\rho_s = .44$) and actuators ($\rho_s = .37$) increase the skill repertoire.
These results suggest that it is easier to implement more skills with a mechanically complex hand, of which anthropomorphic hands are currently the most complex subset. However, there is no notion of task or skill complexity in this analysis. Incorporating a notion of skill difficulty should be addressed in future work.

\subsection{Lack of Sensor Integration in State of the Art}
From our systematic review, we see that the mode of the distribution of the number of sensors used per paper is one, and several publications use none, indicating that sensor data are only sparsely integrated into robotic hand control. Of the 62 papers that include proprioceptive sensing in their setup, only 18 use tactile perception to implement the skills, and 7 of these rely on tactile input only as a binary or average contact signal rather than exploiting its full spatial, temporal, and force resolution. Consequently, our correlation analyses for RQ4 (tactile sensing) and RQ5 (richer sensing), even though showing statistically significant results, may not be meaningful. The integration of haptic information into control pipelines is still underdeveloped.

We have seen rapid progress in robot control recently, with groundbreaking research on the control of complex hands based almost exclusively on visual sensors \cite[e.g.,][]{OpenAI2019Solving,Chen2025ViViDex,Koczy2025Learning}. Integrating rich tactile and proprioceptive feedback with complex hands remains an open challenge, even though human and animal studies show that haptic information is critical for stable grasping, slip detection, and fine force modulation during manipulation \cite{Dahiya2010Tactile,Lederman2009Haptic}. The hardware for high-quality tactile sensing already exists, but current robots rarely use such information beyond simple contact detection.

Robots are still inferior to humans in terms of sensory feedback. We presented examples of complex skills implemented with current robot hands and discussed the case of teleoperation, which suggests that cognition often appears to be the primary constraint rather than hardware and sensing. Surgeons accomplish remarkable tasks with parallel grippers and no explicit haptic feedback, but this ability depends on extensive training and experience to compensate for limited feedback and comes at the cost of increased cognitive load \cite{Odoh2024Performance}. This contrast raises a concrete question for future work: to what extent should advanced robotic hands integrate haptic perception, and how should this feedback be exploited in control architectures?

\subsection{Open Research Questions}

We discovered several interesting research questions that were not the focus of this manuscript or could not be addressed immediately with the available data. We think it is worth investigating these with similar methods in the future.

Birds use a highly flexible neck~\cite{sugasawa2021object} that functions as an equivalent to the human arm during manipulation, and human manipulation relies more on the 3 DoF of the wrist than on DoFs in the hand~\cite{Montagnani2015it}.
Hence, it may be more effective to mimic the flexibility of the human arm with seven DoF, which allows obstacle or torque limits avoidance, than all DoF of the human hand.
\vspace{-9pt}
\begin{mdframed}[backgroundcolor=gray!20, linewidth=0pt]
How important is an anthropomorphic arm design for the manipulation skill repertoire?
\end{mdframed}

Billard and Kragic \cite{Billard2019Trends} point out that the human thumb creates an asymmetry and constrains the orientation of the hand. They entertain the idea of designing hands with more than one opposing finger to enable more dexterous in-hand manipulation, which would usually require two hands.
There are examples of such hands with two pairs of opposing fingers \citep{Morgan2022Complex,Liarokapis2019Combining}.
\vspace{-9pt}
\begin{mdframed}[backgroundcolor=gray!20, linewidth=0pt]
Are the human finger arrangement and thumb a good hand design?
\end{mdframed}

As the research of Sun et al.~\cite{Sun2022MultiObject} on multi-object grasps suggests, more fingers allow handling multiple objects simultaneously and conforming better with the shape of multiple objects. The more tasks have to be solved in parallel, the more fingers are required. However, we only included one article with six fingers.
\vspace{-9pt}
\begin{mdframed}[backgroundcolor=gray!20, linewidth=0pt]
Can more than five fingers solve more manipulation tasks or individual tasks more effectively than the human hand?
\end{mdframed}

Three fingers appear to be a useful sweet spot:
they are sufficient for most human grasps~\cite{Abbasi2016Grasp}, even ten of twelve human grasp types for multi-object grasping can be replicated with a BarrettHand~\cite{Sun2022MultiObject}, they have an increased in-hand manipulation workspace compared to more fingers in a human finger layout~\cite{Feix2021Effect}, a BarrettHand can perform functional tool-use grasps~\cite{Zhu2023HumanLike,Zhu2021HumanLike}, and opposing one finger against two with abduction provides sufficient support for deformable object tasks (picking, folding, and tracing cloth) that would challenge a two-finger gripper~\cite{Donaire2020Versatile}.
\vspace{-9pt}
\begin{mdframed}[backgroundcolor=gray!20, linewidth=0pt]
Is a hand design with three fingers a good trade-off between mechanical simplicity and versatility?
\end{mdframed}

Human mechanoreceptors exhibit high spatial acuity, e.g., Merkel cells with approximately 0.5 mm resolution and Meissner corpuscles with $3-4$ mm, and are tuned to specific frequency ranges, while both cutaneous and proprioceptive afferents conduct rapidly, which together support fine texture discrimination, slip detection, precise force scaling, and accurate timing across the phases of grasping and object manipulation \citep{Dahiya2010Tactile,Liu2020Bioinspired,Lederman2009Haptic,Siegel2006Essential}.
On the other hand, when tactile feedback is lacking, grasp phases become longer, grip-force timing deteriorates, and grasps become unstable even if vision is available. Similarly, impairments in proprioception lead to inaccurate limb positioning and disturbed movement trajectories, making precise manipulation difficult despite intact muscles and vision, showing that loss or severe degradation of proprioceptive resolution limits the range and quality of reachable movements and coordinated actions \citep{Ghez1995Impairments,Lederman2009Haptic}.
\vspace{-9pt}
\begin{mdframed}[backgroundcolor=gray!20, linewidth=0pt]
What spatial, temporal, and force resolution of proprioceptive and tactile sensing is needed for robotic hands?
\end{mdframed}

Teleoperation suggests that a limiting factor may be the cognitive ability of robotic systems, as surgeons can perform complex procedures using simple grippers and tools without haptic feedback. Nonetheless, haptic feedback, even in its current state of development, appears to have a measurable effect on metrics such as applied forces, completion time, accuracy, and success rates \cite{Bergholz2023Benefits}.
\vspace{-9pt}
\begin{mdframed}[backgroundcolor=gray!20, linewidth=0pt]
What is the effect on computation demand for robotics systems with visual-only or visuo-haptic feedback?
\end{mdframed}
\vskip -0.8cm
\begin{mdframed}[backgroundcolor=gray!20, linewidth=0pt]
What skills can only be performed with haptic feedback?
\end{mdframed}

\section{Conclusion}
For prosthetics, emulating the mechanical characteristics of the human hand is a sensible goal. For robots, however, this argument is less applicable and may even be counterproductive in human-robot interaction because of the uncanny valley~\cite{Mori2012Uncanny}.

Are roboticists actually looking at the right defining characteristics of human hands?
The most advanced robotic hands can match the human hand in number of fingers, DoF, and actuators.
However, our results on in-hand manipulation tasks indicate that intelligent manipulation strategies and mechanism designs may be more important for dexterous in-hand manipulation than replicating the appearance of the human hand.
Its key advantage may not lie in its form, but in functional properties of the human hand that current robotic hands have not yet replicated:
sensorimotor control and learning that enable robust skill execution under uncertainty and actuation noise.
Mechanical robustness of the human hand due to its softness, deformability, and passive range of motion enables robust, multimodal sensorimotor control based on motor learning under contacts with the environment.
Hence, function-based biomimicry should be preferred over form-based biomimicry.

There is an advantage of more complex mechanisms when looking at the skill repertoire size though, and the most complex hands are currently anthropomorphic.
No paper considers hands with more than 22 DoF or actuators, and only one paper considers a hand with more than five fingers. Nevertheless, we believe that robots most likely can go beyond human manipulation abilities with more complex designs. These non-anthropomorphic hands should be able to solve complex manipulation problems more effectively or multiple manipulation problems simultaneously, since they can directly control many DoFs of one or multiple objects without repositioning. How this can be done remains an open question.

Our systematic review reveals a key shortcoming of the current state of the art: most hands are tested within only a few categories of Dollar's taxonomy \cite{Dollar2014Classifying}, and there is limited variety even within these. This lack of comprehensive evaluation is especially evident in cases of complex hands, which are seldom tested for in-hand manipulation. To address this, the field needs robust evaluation protocols that can assess complete mechanism-sensor-software systems across all skill categories and complexity levels. Such protocols must demonstrate a hand's functional capabilities, especially by requiring the intelligent use of advanced sensors. For example, to truly evaluate the benefits of tactile sensors, protocols should include complex manipulation tasks that demand fine motor control. Establishing these benchmarks is crucial for making meaningful comparisons of hand designs and for guiding progress in robotic hand development.

\backmatter

\section*{Declarations}

\bmhead{Funding} This work was supported by the European Commission under the Horizon 2020 framework program for Research and Innovation via the APRIL project (project number: 870142) and by the Vibro-Sense Project (project number: 03DPS1242A) funded by the Bundesministerium Forschung, Technologie und Raumfahrt (BMFTR) under the DATIpilot program. This work was partially supported by the German Federal Ministry of Research, Technology and Space (BMFTR) under the Robotics Institute Germany (RIG). Open Access funding provided by the Projekt DEAL (Open access agreement for Germany).
\bmhead{Conflict of interest/Competing interests} The authors declare that they have no conflict of interest.
\bmhead{Ethics approval} This article does not contain any studies with human participants or animals performed by any of the authors.

\bmhead{Open Access} This article is licensed under a Creative Commons Attribution 4.0 International License, which permits use, sharing, adaptation, distribution and reproduction in any medium or format, as long as you give appropriate credit to the original author(s) and the source, provide a link to the Creative Commons licence, and indicate if changes were made. The images or other third party material in this article are included in the article's Creative Commons licence, unless indicated otherwise in a credit line to the material. If material is not included in the article's Creative Commons licence and your intended use is not permitted by statutory regulation or exceeds the permitted use, you will need to obtain permission directly from the copyright holder. To view a copy of this licence, visit \url{http://creativecommons.org/licenses/by/4.0/}.

\bmhead{Consent to participate} Not applicable
\bmhead{Consent for publication/Informed consent} Not applicable
\bmhead{Availability of data and materials} Overviews of the analyzed robotic hands and the papers considered for the systematic review as well as the code are available at \url{https://github.com/AlexanderFabisch/JIRS-HandsSurvey}.
\bmhead{Code availability} The analysis code is available at \url{https://github.com/AlexanderFabisch/grasp_survey_scripts}.

\bmhead{Author contribution statement}
All authors (A.F., W.Z.E.A., C.S., N.N.G.) defined the concept of the manuscript, defined the research questions for the systematic review, annotated data for the systematic review, wrote the discussion and conclusion, and edited and reviewed the manuscript.
A.F. wrote Section \ref{sec:humanHand_biomechanics} and is responsible for the statistical approach in the systematic review.
N.N.G. wrote Sections \ref{sec:human_perception} and \ref{sec:robot_perception}.
W.Z.E.A. wrote Section \ref{sec:robotic_hands_mechanics}.

\begin{appendices}

\section{Muscles of the Human Hand}
\label{app:human_hand_muscles}

\paragraph{Wrist (extrinsic muscles, Fig.~\ref{fig:muscles-forearm})}
Several muscles work together to flex and extend the wrist joint.\footnote{\textit{Flexor carpi radialis}, \textit{flexor carpi ulnaris}, and \textit{palmaris longus} flex the wrist joint. \textit{Extensor carpi ulnaris}, \textit{extensor carpi radialis longus}, and \textit{extensor carpi radialis brevis} extend the wrist joint \citep{Young2013Anatomy}.} However, most of these muscles also have another function. All \textit{carpi radialis} muscles contribute to abduction, and all \textit{carpi ulnaris} muscles contribute to adduction of the wrist. Furthermore, there are extrinsic finger muscles that contribute to wrist flexion and extension \citep{Young2013Anatomy}.

\paragraph{Extrinsic finger muscles, excluding the thumb (Fig.~\ref{fig:muscles-forearm})}
The extrinsic flexors and extensors of the fingers act on multiple joints, including the wrist. Most notably are
\begin{itemize}[leftmargin=*,noitemsep,topsep=0pt]
\item \textit{Flexor digitorum superficialis} (FDS): its four tendons attach to the middle phalanges 2--5 and it flexes the fingers and the wrist.
\item \textit{Flexor digitorum profundus} (FDP): its four tendons attach to the distal phalanges 2--5, passing through the tendons of FDS, and it flexes the fingers and the wrist.
\item \textit{Extensor digitorum communis} (EDC): its four tendons attach to the distal phalanges 2--5 and it extends the fingers and the wrist.
\end{itemize}
Furthermore, the extrinsic finger muscles \textit{extensor indicis} and \textit{extensor digiti minimi} extend all joints of only one finger---index and little finger, respectively---and the wrist.

\paragraph{Intrinsic and extrinsic muscles of the thumb (Fig.~\ref{fig:muscles-hand})}
The thumb is an unusual yet essential finger with its large range of motion in the TMC joint and ability to oppose the other fingers. Eight muscles control it. The two DoF of the TMC joint are controlled by all intrinsic\footnote{Intrinsic muscles that control the thumb: \textit{opponens pollicis}, abductor \textit{pollicis brevis}, flexor \textit{pollicis brevis}, adductor \textit{pollicis}.} and extrinsic\footnote{Extrinsic muscles that control the thumb: flexor \textit{pollicis longus}, extensor \textit{pollicis longus}, extensor \textit{pollicis brevis}, abductor \textit{pollicis longus}.} muscles of the thumb. The MCP joint is flexed by flexor \textit{pollicis longus} and \textit{brevis} and extended by extensor \textit{pollicis longus} and \textit{brevis}. The IP joint is only flexed by flexor \textit{pollicis longus} and extended by extensor \textit{pollicis longus}. Abductor \textit{pollicis longus} also contributes to wrist abduction, and flexor \textit{pollicis longus} to wrist flexion.

\paragraph{Other intrinsic muscles (Fig.~\ref{fig:muscles-hand})}
Four \textit{dorsal interossei} muscles and three \textit{palmar interossei} muscles perform abduction and adduction of fingers 2--5. They also assist the \textit{lumbricals}, which are located between tendons of the FDP at the palm and the back side of the distal phalanges 2--5, in their function. They flex the MCP joints and extend the PIP and DIP joints, allowing for individual flexion of the MCP joints.

There are also specific muscles that have a single function and can be found only in one or two fingers.
Abductor \textit{digiti minimi}: it abducts the little finger so that the abduction of the little finger can be stronger than that of fingers 2--4.
Flexor \textit{digiti minimi brevis}: flexes the little finger at the MCP joint.
Opponens \textit{digiti minimi}: brings the little finger into opposition, much less than the thumb, though, and also hollows the palm.
\textit{Palmaris brevis} does not seem to be functionally relevant, but provides protection to underlying tissue \citep{Moore2017Structural}.

\end{appendices}

\bibliography{references}

\end{document}